\documentclass[a4paper,fleqn]{cas-dc}

\usepackage[numbers, sort&compress]{natbib}
\usepackage{amsfonts} 
\usepackage{nicefrac}
\usepackage{microtype}  
\usepackage{graphicx}
\usepackage{colortbl}
\usepackage{amssymb}
\usepackage{gensymb}
\usepackage{tabularx}
\usepackage{float}
\usepackage{amsthm}
\usepackage{amsmath}
\usepackage[pagewise,switch]{lineno}
\usepackage[ruled,vlined,linesnumbered]{algorithm2e}
\usepackage{caption}
\usepackage{subcaption}

\begin{document}
\let\WriteBookmarks\relax
\def\floatpagepagefraction{1}
\def\textpagefraction{.001}
\shorttitle{}
\shortauthors{Khalifa et~al.}

\title [mode = title]{A Review of Hidden Markov Models and Recurrent Neural Networks for Event Detection and Localization in Biomedical Signals}

\author[1]{Yassin Khalifa}

\address[1]{Department of Electrical and Computer Engineering, University of Pittsburgh, Pittsburgh, PA, USA}

\author[2]{Danilo Mandic}

\address[2]{Department of Electrical and Computer Engineering, Imperial College, London, SW7 2BT United Kingdom}

\author[1,3,4,5]{Ervin Sejdi\'{c}}[orcid=0000-0003-4987-8298]
\cormark[1]
\ead{esejdic@ieee.org}
\ead[URL]{www.imedlab.org}

\address[3]{Department of Bioengineering, University of Pittsburgh, Pittsburgh, PA, USA}
\address[4]{Department of Biomedical Informatics, University of Pittsburgh, Pittsburgh, PA, USA}
\address[5]{Intelligent Systems Program, University of Pittsburgh, Pittsburgh, PA, USA}

\cortext[cor1]{Corresponding author}

\begin{abstract}
Biomedical signals carry signature rhythms of complex physiological processes that control our daily bodily activity. The properties of these rhythms indicate the nature of interaction dynamics among physiological processes that maintain a homeostasis. Abnormalities associated with diseases or disorders usually appear as disruptions in the structure of the rhythms which makes isolating these rhythms and the ability to differentiate between them, indispensable. Computer aided diagnosis systems are ubiquitous nowadays in almost every medical facility and more closely in wearable technology, and rhythm or event detection is the first of many intelligent steps that they perform. How these rhythms are isolated? How to develop a model that can describe the transition between processes in time? Many methods exist in the literature that address these questions and perform the decoding of biomedical signals into separate rhythms. In here, we demystify the most effective methods that are used for detection and isolation of rhythms or events in time series and highlight the way in which they were applied to different biomedical signals and how they contribute to information fusion. The key strengths and limitations of these methods are also discussed as well as the challenges encountered with application in biomedical signals.
\end{abstract}


\maketitle

\section{Introduction}
	\label{S:Introduction}
	Physiological processes are complex tasks performed by the different systems of the human body in a rarely periodic but rather irregular manner to deliver an action that could be biochemical, electrical, or mechanical \cite{glass_synchronization_2001, rangayyan_biomedical_2002}. Some of these actions are obvious like heart beating, breathing, and other physical activities and some are not as obvious like hormonal stimulation that regulates multiple body functions. The action produced can be usually manifested as some sort of a signal that holds information about the parent physiological process \cite{rangayyan_biomedical_2002}. Disruptions in these physiological processes associated with diseases, lead to the development of pathological processes that alter the performance of the human body. Both normal and pathological processes in addition to other artifacts from the environment and surrounding processes, are all held in the manifested signals and the associated changes in their waveform. These signals are called biomedical signals and can be of many forms including the electrical form (potential or current changes) or physical (force or temperature) \cite{rangayyan_biomedical_2002}.\par
	
	Artificial intelligence is currently taking over to empower a variety of assistive technologies that help solve the problems of the healthcare sector given the continuously increasing cost and shortage of professional caregivers. These technologies are advancing to perform not only diagnosis but also intervention and curing due to the superior sensitivity, adaptability, and fast response. Of these assistive technologies, computer aided diagnosis and wearable systems are powered by the virtual side of artificial intelligence (machine learning techniques) and play a vital role in anomaly detection, monitoring, and even emergency response \cite{rashidi_survey_2013}. The rise of such systems has led to the evolution of biomedical signal analysis which has been the focus of researchers for the last couple of decades. This evolution not only included the macro-analysis of gross processes but also the detection and analysis of micro-events within each gross process \cite{rashidi_survey_2013}. As mentioned before, biomedical signals carry the signatures of many processes and artifacts, which makes the extraction/identification of the specific part of interest (called event or epoch), the first step of any systematic signal analysis or monitoring \cite{kim_biomedical_2009}. Further, the need for robust event extraction algorithms for biomedical signals is driven by the exponential growth of the amount and complexity of data generated by biomedical systems \cite{andreu-perez_big_2015}. Moreover, reducing the human-dependent steps in the analysis, mitigates the reliability and subjectivity issues associated with human tolerance. \par
	
	Epoch extraction is not only essential for systematic signal analysis, but also substantial to information fusion for multi-channel systems and/or sensor networks which represent a large portion of biomedical-signal-based decision-making systems nowadays. Multiple fusion models can employ epoch extraction and event detection to overcome different obstacles including but not limited to signal synchronization and feature fusion \cite{gravina_multi-sensor_2017,mandic_data_2005}. In complementary data-level fusion, events can be used to align signals as preparation for feature extraction such as using heart beats to align the signals from multiple electrocardiography (ECG) leads. In feature-level fusion models, event detection can be used to combine features from different signals during only the events of interest that contribute to morphology analysis and the decision-making process \cite{mandic_data_2005,mandic_signal_2008}.\par
	
	Epoch extraction algorithms have been used repeatedly in segmentation of many biomedical signals, including, but not limited to, heart sound and ECG \cite{huiying_heart_1997, pan_real-time_1985}, electroencephalography (EEG) \cite{srinivasan_approximate_2007, kannathal_entropies_2005, schlogl_characterization_2005}, and swallowing vibrations \cite{khalifa_non-invasive_2020, khalifa_upper_2020, sejdic_segmentation_2009, damouras_online_2010}. Such algorithms immensely depend on modeling time-series, the paradigm that is not explicitly provided by regular machine learning and sequence-agnostic models such as support vector machines, regression, and feed forward neural networks \cite{lipton_critical_2015}. These models depend on a major assumption that the training and test examples are independent and not related in time or space which in result initiates a reset to the entire state of the model \cite{lipton_critical_2015}. Particularly speaking, splitting time series into data chunks and using consecutive chunks independently in building models is unacceptable because even in the case of modeling a time series with iid processes, the underlying processes might be longer than a single chunk which induces dependency between consecutive chunks. \par
	
	Sliding window approach has been introduced to tackle the problem of dependence between consecutive chunks through using an overlap which guarantees that a part of each chunk will be carried over to the next chunk. Although this might be useful in modeling many processes, it fails to model long range dependencies and requires the optimization of both data chunk and overlap lengths to best represent the target processes. Additionally, using windowing in time domain provokes a sort of distortion to the frequency representation due to the leakage effect and can only be used for modeling fixed-length input/output scenarios \cite{lipton_critical_2015}. All of this raised the need for models capable of selectively transferring states across time, processing sequences of not necessarily independent elements, and yielding a computational paradigm that can handle variable-length inputs and outputs \cite{hochreiter_long_1997}. It was not that long before the researchers started to bring stochastic-based models \cite{rabiner_tutorial_1989} and design deep recurrent networks \cite{hochreiter_long_1997} to perfectly fit the event extraction problems and overcome the limitations of regular machine learning methodologies. \par
	
	Multiple models have been offered for time dependency representation including Hidden Markov models (HMMs) and Recurrent Neural Networks (RNNs). HMMs were introduced as an extension to Markov chains to probabilistically model a sequence of observations based on an unobserved sequence of states \cite{rabiner_tutorial_1989}. On the other hand, RNNs generalize the feed-forward neural networks with the ability to process sequential data one step at a time while selectively transferring information across sequence elements \cite{lipton_critical_2015}. Hence, RNNs are successful in modeling sequences with unknown length, components that are not independent, and multi-scale sequential dependencies \cite{hochreiter_long_1997, werbos_backpropagation_1990, rumelhart_learning_1986}. Further, RNNs overcame a major HMM limitation in modeling the long-range dependencies within the sequences \cite{lipton_critical_2015, graves_neural_2014}.\par
	
	In this manuscript, we review the fundamental methods developed for event extraction in biomedical signals and unravel the key differences between these methods based on the state-of-the-art practices and results. We show the theoretical and practical aspects for most of the methods and the way in which they were used to handle the time modeling in event detection problems. Further, we discuss the recent major machine learning applications in biomedical signal processing and the anticipated advances for future implementations. \par

\section{Hidden Markov Models}
	\label{S:HMMs}
	A time series can be characterized using either deterministic or stochastic models. Deterministic models usually describe the series using some specific properties such as being the sum of sinusoids or exponentials and aim to estimate the values of the parameters contributing to these properties (e.g. amplitude, frequency, and phase of the sinusoids) \cite{rabiner_tutorial_1989}. On the other hand, statistical models assume that the series can be described through a parametric random process whose parameters can be estimated in a well-defined way \cite{rabiner_tutorial_1989, cohen_hidden_1998}. HMMs belong to the category of statistical models and usually are referred to as probabilistic functions of Markov chains in the literature \cite{rabiner_tutorial_1989, baum_statistical_1966}. \par

	\subsection{Markov Chains}
		\label{SS:MC}
		Markov chain is a stochastic process modeled by a finite state machine that can be described at any instance of time to be one of $N$ distinct states. These states can be tags or symbols representing the problem of interest. The machine may stay at the same state or switch to another state at regularly spaced discrete times according to  a set of transition probabilities associated with each state \cite{rabiner_tutorial_1989, cohen_hidden_1998} and the transition probabilities are assumed to be time independent. The initial state is deemed to be known and the transition probabilities are described using the transition matrix: $A = \{a_{ij}\};$ where $a_{ij}$ is the transition probability from state $S_i$ to state $S_j$ and both $i$, and $j$ can take values from $1$ to $N$. The actual state at time $t$ is denoted as $q_t$ and for a full description of the probabilistic model, the current state as well as at least the state previous to it (for a first order Markov chain), need to be specified. The first order Markov chain assumes that the current state depends only on the previous state: $P(q_t=j|q_{t-1}=i, q_{t-2}=k, \dots) = P(q_t=j|q_{t-1}=i)$. This results in the following properties for the transition probabilities:
		
		\begin{eqnarray*}
			&a_{ij} &= P(q_t=j|q_{t-1}=i); \ \ \ \ \ \ \ i\geq 1,\ j\leq N\\
			&a_{ij}&\geq 0\\
			&\sum\limits_{j=1}^{N}a_{ij} &= 1
		\end{eqnarray*}
		
		The probability of being at state $S_i$ at $t = 1$ is denoted as $\pi_i$, and the initial probability distribution as:
		
		\begin{eqnarray*}
			\pi_i &=& P[q_1 = S_i]; \ \ \ \ \ \ \ 1\leq i\leq N\\
			\Pi &=& [\pi_1, \pi_2, \dots, \pi_N ]^T
		\end{eqnarray*}
		
\begin{figure}[!h]
			\centering
			\includegraphics[width=0.8\columnwidth]{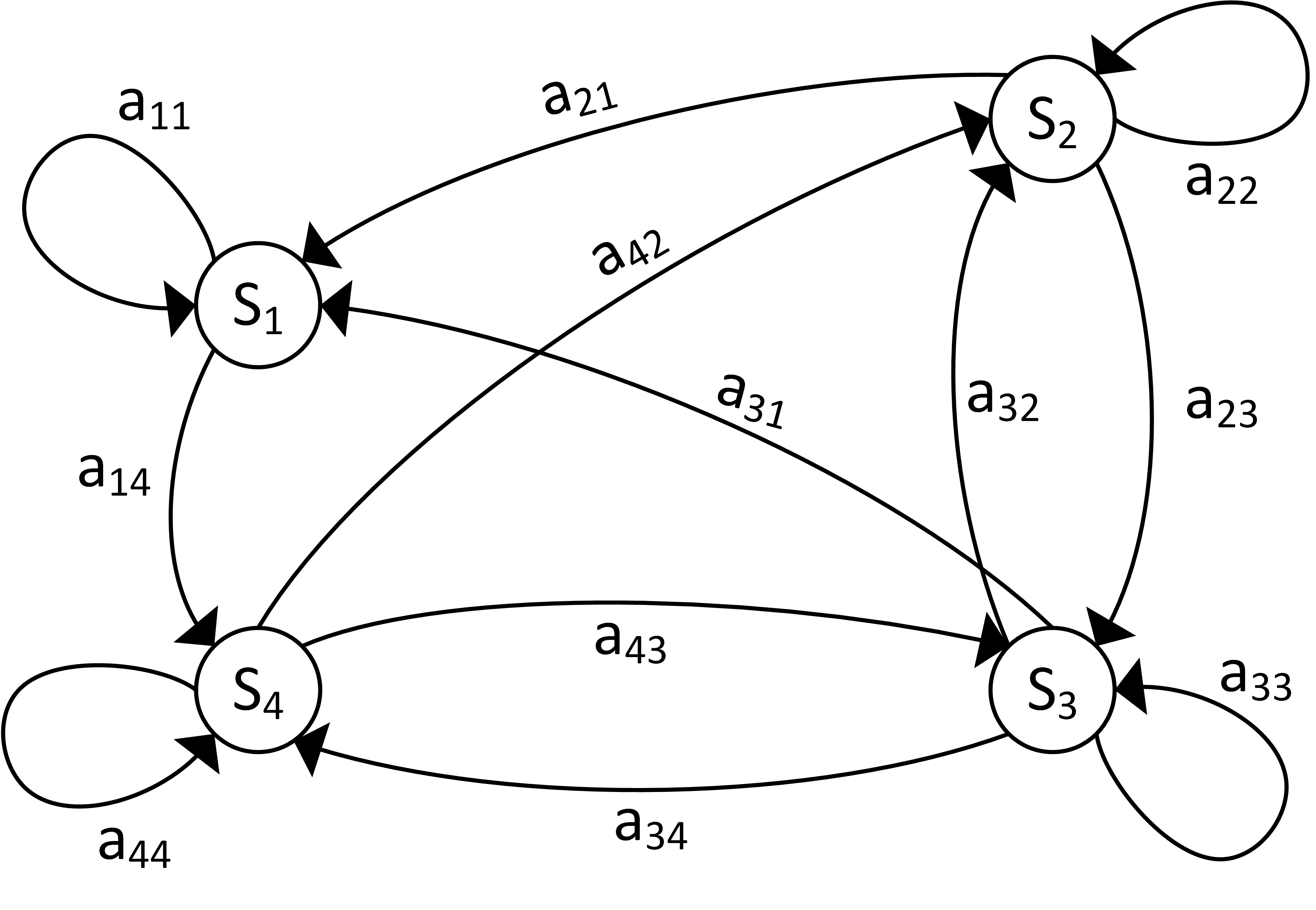}\\
			\caption{An example of a Markov chain with 4 states, $S_1$ to $S_4$, and selected state transitions. A set of probabilities is associated with each state to indicate how the system undergoes state change from one state to itself or another at regular discrete times.}
			\label{fig:markonchex}
		\end{figure}

		An example of a 4-states Markov chain is shown in Fig. \ref{fig:markonchex}. This stochastic process is called the observable Markov model since each state corresponds to a visible (observable) event.
		
	\subsection{Hidden Markov Models}
		\label{SS:HMMs}
		So far, we introduced Markov chains in which each state corresponds to an observable event, however this is insufficient for most of the applications where the states cannot always be observable.  Therefore, Markov chain models are extended to HMMs which can be widely used in many applications \cite{rabiner_tutorial_1989}. HMM is considered a doubly stochastic process with one of them hidden or not observable; states, in this case, are hidden from the observer \cite{rabiner_tutorial_1989}. An HMM is characterized through the following properties \cite{rabiner_tutorial_1989, cohen_hidden_1998}:
		
		\begin{enumerate}
			\item The number of states, $N$, included in the model. As mentioned before, the states are usually hidden in HMMs but sometimes they have a physical significance.
			\begin{equation*}
			q_t \in \{S_1, S_2, \dots, S_N\}
			\end{equation*}
			\item The number of distinct observations, a state can take, $M$.
			\item The state transition matrix or distribution $A=\{a_{ij}\}$.
			\item The observation probability distribution for each state $B = \{b_j(k)\} = P[v_k\ at\ t | q_t = S_j]$; where $v_k$ represents an element of the distinct observations that a state can take and $1\leq j\leq N,\ 1\leq k\leq M$.
			\item The initial state distribution $\Pi = \{\pi_i\}$.
		\end{enumerate}
		
		When known, the previously mentioned parameters can be used to fully describe the HMM ($\lambda(A,\ B,\ \Pi)$) and generate an observation sequence $O = \{O_1, O_2, \dots, O_T\}$ as in the algorithm shown in Algorithm \ref{alg:ob_gen}.
		
		\begin{algorithm}[!h]
			\begin{footnotesize}
				\SetAlgoLined
				Set $t\ =\ 1$\;
				Choose an initial state $q_1\ =\ S_i$ according to $\Pi$\;
				\While{$t\ \le T$}{
					Choose $O_t = v_k$ according to the observation distribution in the current state ($b_i(k)$)\;
					Move from the current state $S_i$ to the new state $q_{t+1} = S_j$ according to $a_{ij}$\;
					set $t\ =\ t\ +\ 1$\;
				}\
				\KwResult{$O = \{O_1, O_2, \dots, O_T\}$}
			\end{footnotesize}
			\caption{HMM as observations generator}
			\label{alg:ob_gen}
		\end{algorithm}
		
		For the model to be useful for trending applications, it must address three fundamental problems \cite{jurafsky_speech_2009}:
		\begin{itemize}
			\item \textbf{Likelihood:} Computing the probability of an observation sequence $O = \{O_1, O_2, \dots, O_T\}$, given the model ($P(O|\lambda)$).
			\item \textbf{Decoding:} Choosing the optimal hidden state sequence $Q = \{q_1, q_2, \dots, q_T\}$ that best represents a given observation sequence ($O = \{O_1, O_2, \dots, O_T\}$).
			\item \textbf{Estimation:} Adjusting the model parameters $\lambda(A,\ B,\ \Pi)$ to maximize the likelihood of a given sequence of observations $O$.
		\end{itemize}

	\subsection{Likelihood Problem Solution}	
		\label{SS:LIKE}
		In the case of Markov chains, where the states are not hidden, the computation of the likelihood is much easier as it narrows the computational burden to just multiplying the transition probabilities within the underlying state sequence. In HMMs, states are hidden which necessitates including all possible state sequences in computing the joint probability ($N^T$ possible hidden state sequences). A dynamic programming solution called the forward-backward algorithm was created for the likelihood problem with a simple time complexity \cite{rabiner_tutorial_1989}. The forward-backward algorithm sums the probabilities of all possible state sequences that could be included in generating the target observation sequence. The algorithm considers an efficient way to calculate the probability through defining and inductively computing the forward variable $\alpha(t, i)$ which represents the probability of the partial observation sequence $P(O_1\ O_2\ \dots\ O_t, q_t=S_i|\lambda)$ \cite{baum_inequality_1967, baum_growth_1968, rabiner_tutorial_1989}. The forward algorithm for the likelihood problem is fully described as follows:
		
		\begin{algorithm}[!h]
			\begin{footnotesize}
				\SetAlgoLined
				$O = \{O_1, O_2, \dots, O_T\}$\;
				$S\ \in\ \{S_1, S_2, \dots, S_N\}$\;
				Create the forward probability table $\alpha[T, N]$\;
				\ForEach{state $S\ \in\ \{S_1, S_2, \dots, S_N\}$}{
					$\alpha[1, S] \gets \pi_S \times b_S(O_1)$\tcp*{Initialization}
				}\
				\ForEach{time step $t\ \in\ {2, 3, \dots, T}$}{
					\ForEach{state $S\ \in\ \{S_1, S_2, \dots, S_N\}$}{
						$\alpha[t, S]\gets \sum\limits_{\hat{S} = S_1}^{S_N}\alpha[t-1, \hat{S}]\times a_{\hat{S}, S}\times b_S(O_t)$\tcp*{Induction}
					}
				}\
				$P(O|\lambda(A,\ B,\ \Pi))\gets \sum\limits_{S=S_1}^{S_N}\alpha[T, S]$\tcp*{Termination}\
				\KwResult{$P(O|\lambda(A,\ B,\ \Pi))$}
			\end{footnotesize}
			\caption{The forward algorithm}
			\label{alg:frwrdbkwrdAlg}
		\end{algorithm}
		
		As a part of the forward-backward algorithm, another variable is considered that will be of help in the solution of the estimation problem. The variable is called the backward probability table, $\beta(t, i) = P(O_{t+1},\ O_{t+2},\ \dots,\ O_{T}|q_t=S_i, \lambda(A,\ B,\ \Pi))$, which represents the probability of the partial observation sequence that starts one time step after the current observation, given the current state $S_i$ and the model. The backward probability can be calculated in a similar way as the forward probability (Algorithm \ref{alg:bckwrdAlg}).
		
		\begin{algorithm}[!h]
			\begin{footnotesize}
				\SetAlgoLined
				Create the backward probability table $\beta[T, N]$\;
				\ForEach{state $S\ \in\ \{S_1, S_2, \dots, S_N\}$}{
					$\beta[T, S] \gets 1$\tcp*{Initialization}
				}\
				\ForEach{time step $t\ \in\ {T-1, T-2, \dots, 1}$}{
					\ForEach{state $S\ \in\ \{S_1, S_2, \dots, S_N\}$}{
						$\beta[t, S]\gets \sum\limits_{\hat{S} = S_1}^{S_N}\beta[t+1, \hat{S}]\times a_{S, \hat{S}}\times b_{\hat{S}}(O_{t+1})$\tcp*{Induction}
					}
				}\
				\KwResult{$\beta[T, N]$}
			\end{footnotesize}
			\caption{Computing the backward probability}
			\label{alg:bckwrdAlg}
		\end{algorithm}

	\subsection{Decoding Problem Solution: The Viterbi Algorithm}	
		\label{SS:DEC}
		Finding the optimal hidden states sequence that best represents a sequence of observations is more challenging compared to the likelihood problem. Unlike the likelihood problem, the decoding problem does not have an exact solution unless the model is degenerate, which makes it hard to choose the optimality criterion that judges the state sequence \cite{rabiner_tutorial_1989}. For example, one may choose states based on the individual likelihood of occurrence which achieves the maximum number of correct states individually but not for the overall computed sequence \cite{rabiner_tutorial_1989}. Another way to solve the decoding problem can be achieved through running the forward-backward algorithm for all possible hidden state sequences and choose the sequence with the maximum likelihood probability, however this is computationally unfeasible \cite{jurafsky_speech_2009}.\par
		
		In the same way as the forward-backward algorithm, the Viterbi algorithm solves the decoding problem using dynamic programming. The algorithm recursively computes the probability of being in a state $S_j$ at time $t$ taking in consideration the most probable state sequence (path) $q_1,\ q_2,\ \dots,\ q_{t-1}$ that leads to this state. The Viterbi algorithm is shown in Algorithm \ref{alg:VitAlg}.
		
		\begin{algorithm}[!h]
			\begin{footnotesize}
				\SetAlgoLined
				$O = \{O_1, O_2, \dots, O_T\}$\;
				$S\ \in\ \{S_1, S_2, \dots, S_N\}$\;
				Create the best path probability table $\delta[T, N]$\;
				Create the state index table (the index of state that by adding to the path, maximizes $\delta$) $\psi[T, N]$\;
				\ForEach{state $S\ \in\ \{S_1, S_2, \dots, S_N\}$}{
					$\delta[1, S] \gets \pi_S \times b_S(O_1)$\tcp*{Initialization}\
					$\psi[1, S]\gets 0$\;
				}\
				\ForEach{time step $t\ \in\ {2, 3, \dots, T}$}{
					\ForEach{state $S\ \in\ \{S_1, S_2, \dots, S_N\}$}{
						$\delta[t, S]\gets \max\limits_{\hat{S}=S_1}^{S_N}\delta[t-1, \hat{S}]\times a_{\hat{S}, S} \times b_S{O_t}$\tcp*{Induction}\
						$\psi[t, S]\gets \arg\max\limits_{\hat{S}=S_1}^{S_N}\delta[t-1, \hat{S}]\times a_{\hat{S}, S} \times b_S{O_t}$\;
					}
				}\
				$P^*\gets \max\limits_{S=S_1}^{S_N}\delta[T, S]$\tcp*{Termination}\
				$q_T^*\gets \arg\max\limits_{S=S_1}^{S_N}\delta[T, S]$\;
				\For{$t\ \in\ \{T,\ T-1,\ T-2,\ \dots,\ 2\}$}{
					$q_{t-1}^*\gets \psi[t, q_t]$\tcp*{Backtracking}
				}\
				\KwResult{The optimal state sequence: $q_{1}^*,\ q_{2}^*,\ \dots,\ q_{T}^*$}
			\end{footnotesize}
			\caption{The Viterbi algorithm}
			\label{alg:VitAlg}
		\end{algorithm}

	\subsection{Model Estimation Problem Solution}	
		\label{SS:MEST}
		
		The third problem can be formulated as finding HMM's model parameters $(A, B, \Pi)$ to maximize the conditional probability of observation sequence, given that model \cite{rabiner_tutorial_1989}. Such a problem doesn't have an analytical solution, however, iterative methods can be used to find a local maxima for $P(O|\lambda)$. Here, we focus on the Baum-Welch algorithm that is based on the expectation-maximization method \cite{baum_inequality_1972, dempster_maximum_1977}. The algorithm is based on maximizing Baum's auxiliary function over the updated model parameters $\lambda$,
		
		\begin{equation*}
			Q(\bar{\lambda},\ \lambda) = \sum\limits_{\forall q}^{} P(O_{1:T},q_{1:T}|\bar{\lambda}) \log P(O_{1:T},q_{1:T}|\lambda),
		\end{equation*}
		where $P(O_{1:T},q_{1:T}|\lambda) = \pi\prod\limits_{t=1}^{T-1} a_{q_t, q_{t+1}} b_{q_{t+1}}(O_{t+1})$, and $\bar{\lambda}$ is the initial model. The iterations are performed based on the calculations by the forward-backward probabilities described previously in the solution of the first two problems, and they go as described in Algorithm \ref{alg:EstAlg}.
		
		\begin{algorithm}[!h]
			\begin{footnotesize}
				\SetAlgoLined
				$O = \{O_1, O_2, \dots, O_T\}$\;
				$S\ \in\ \{S_1, S_2, \dots, S_N\}$\;
				Initialize $\bar{\lambda} = \lambda(A,\ B,\ \Pi)$\;
				\Repeat{Convergence}{
					Using the forward-backward algorithm and $\bar{\lambda}$ calculate $\alpha[T, N]$ and $\beta[T, N]$\;
					Create the probability tables $\xi[T, N, N]$ (the probability of being in a state $S_i$ at time $t$ and a state $S_j$ at time $t+1$) and $\gamma[T, N]$ (the probability of being in a state $S_i$ at time $t$)\;
					\ForEach{time step $t\ \in\ {2, 3, \dots, T}$}{
						\ForEach{state $S\ \in\ \{S_1, S_2, \dots, S_N\}$}{
							\ForEach{state $S^*\ \in\ \{S_1, S_2, \dots, S_N\}$}{
								$\xi[t, S, S^*] \gets \frac{\alpha[t, S]\times a_{S, S^*}\times b_{S^*}(O_{t+1})\times \beta[t+1, S^*]}{\sum\limits_{S=S_1}^{S_N} \sum\limits_{S^*=S_1}^{S_N} \alpha[t, S]\times a_{S, S^*}\times b_{S^*}(O_{t+1})\times \beta[t+1, S^*]}$\;
							}\
							$\gamma[t, S] \gets \sum\limits_{\bar{S}=S_1}^{S_N}\xi[t, S, \bar{S}]$\;
						}
					}\
					$\bar{\pi}_S \gets \gamma[1, S]$\;
					$\bar{a}_{S, S^*} \gets \frac{\sum\limits_{t=1}^{T-1}\xi[t, S, \bar{S}]}{\sum\limits_{t=1}^{T-1}\gamma[t, S]}$\;
					$\bar{b}_{S}(k) \gets \frac{\sum\limits_{\substack{t=1\\ s.t.\ O_t=v_k}}^{T}\gamma[t, S]}{\sum\limits_{t=1}^{T}\gamma[t, S]}$\;
					$\bar{\lambda} = \lambda(\bar{A}, \bar{B}, \bar{\Pi})$\;
				}\
				\KwResult{$\lambda(A,\ B,\ \Pi)$}
			\end{footnotesize}
			\caption{The estimation algorithm}
			\label{alg:EstAlg}
		\end{algorithm}

	\subsection{Continuous Density HMM}
		The previously described adaptations for HMM problems are based on the requirement that the observations are discrete which is considered restrictive because in most cases they are continuous. Therefore, a necessary first step will be the transformation of continuous observation sequence into a discrete vector. This can be done through dividing the observations' space into sub-spaces and using codebooks to give discrete symbol/value for each sub-space \cite{cohen_hidden_1998}; however, this introduces quantization errors into the problem. One way to overcome this, is using continuous observation densities in HMM's. The finite mixture representation of the observation density function, is one of the representations that has a formulated re-estimation procedure: $b_j(O) = \sum\limits_{m=1}^{M} c_{jm} \mathfrak{N}[O, \mu_{jm}, U_{jm}],$ where $1\leq j\leq N$, $O$ is the observation vector, $c_{jm}$ is the mixture coefficient for the $m^{th}$ mixture in state $j$, and $\mathfrak{N}$ is an elliptically or long-concave symmetric density with a mean vector of $\mu_{jm}$ and a covariance matrix of $U_{jm}$ \cite{liporace_maximum-likelihood_1982, juang_maximum-likelihood_1985, levinson_maximum_1986}. A Gaussian density function is usually used for $\mathfrak{N}$; however, other non-Gaussian models have been considered as well in many applications \cite{safont_multichannel_2019, salazar_including_2010}. The pdf is guaranteed to be normalized, given that the mixture coefficients satisfy the following stochastic conditions: $\sum\limits_{m=1}^{M} c_{jm} = 1$ and $c_{jm} \geq 1$, where $1\leq j\leq N,\ 1\leq m\leq M$. The parameters of the observation density function ($c_{jm}, \mu_{jm}, U_{jm}$) can be estimated through the modified Baum-Welch algorithm \cite{rabiner_tutorial_1989}. Using continuous density in HMM makes it more accurate; however, it requires a larger dataset and a more complex algorithm to train.\par

	\subsection{State Duration in HMM}	
\begin{figure}[!h]
			\centering
			\begin{subfigure}{\columnwidth}
				\centering
				\includegraphics[width=0.7\linewidth]{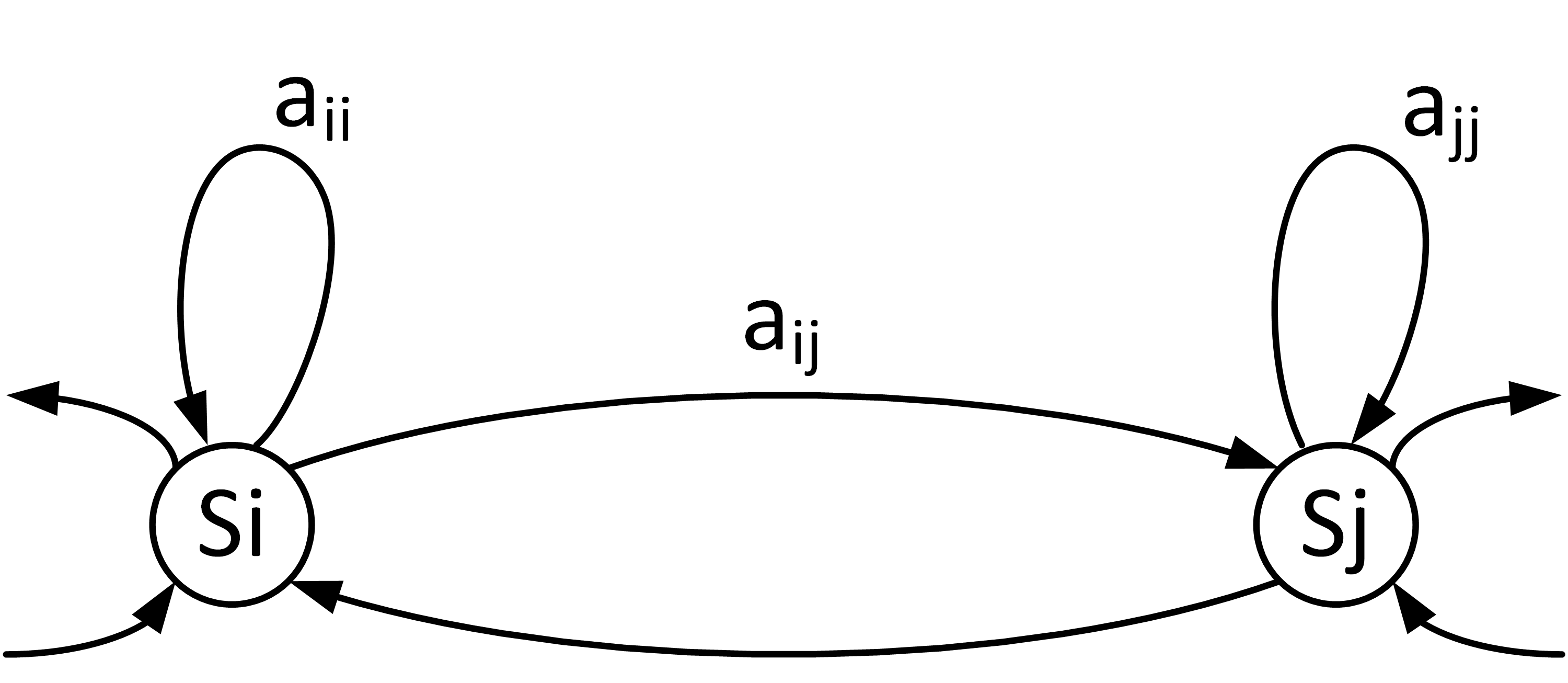}
				\caption{}
				\label{fig:hmmdur_a}
			\end{subfigure}
			\begin{subfigure}{\columnwidth}
				\centering
				\includegraphics[width=0.7\linewidth]{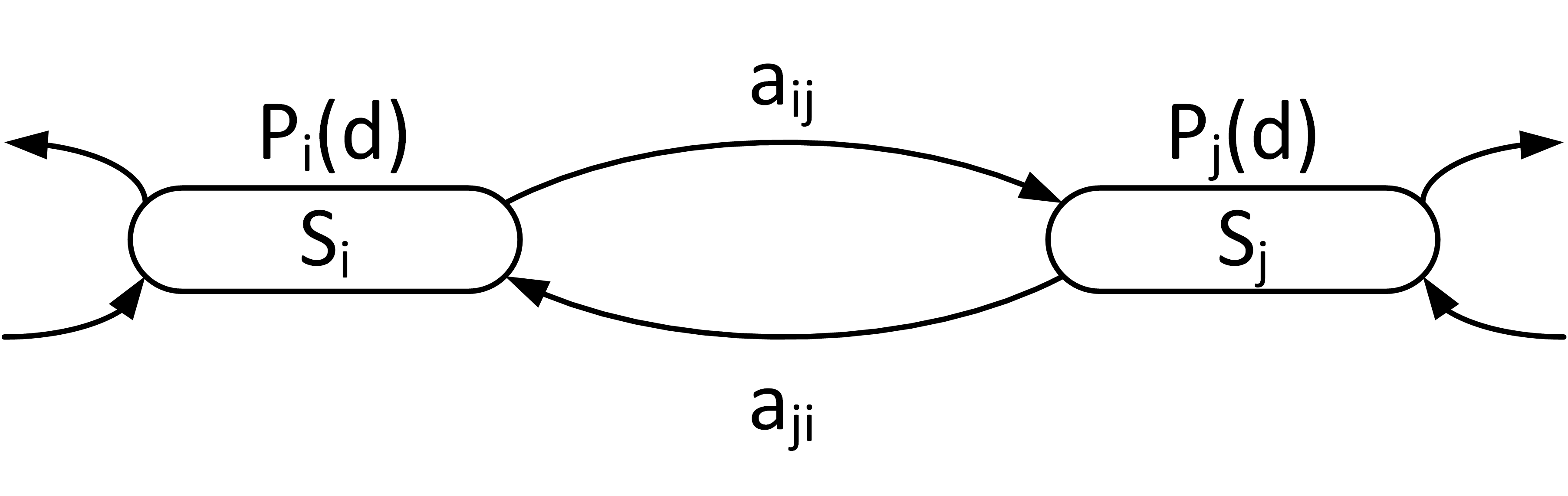}
				\caption{}
				\label{fig:hmmdur_b}
			\end{subfigure}
			\caption{An illustration of interstate connections in HMMs. (a) represents a normal HMM with self transitions from each state back to itself. (b) represents a variable duration HMM with no self state transition and specified state duration densities.}
			\label{fig:hmmdur}
		\end{figure}

		One of the convenient ways to include state duration in HMMs, especially with physical signals, is through explicitly modeling the duration density and setting the self-transition coefficients into zeros \cite{rabiner_tutorial_1989}. The transition from a state to another only occurs after a certain number of observations, specified by duration density, is made in the current state as shown in Fig. \ref{fig:hmmdur}. In normal HMMs, the states have exponential duration densities that depend on the self transition coeeficients $a_{ii}$ and $a_{jj}$ as in Fig. \ref{fig:hmmdur}(a). In HMMs where state duration is modeled by explicit duration densities, there is no self transition and the transition happens only after a specific number of observations determined by the duration density as in Fig. \ref{fig:hmmdur}(b). The re-estimation formulae needed for model estimation can be defined through including the state duration in the calculation of forward and backward variables. The re-estimation formulae can be found in detail in the tutorial of \citet{rabiner_tutorial_1989}.
		
\section{Recurrent Neural Networks}
	Neural networks are biologically-inspired computational models that are composed of a set of artificial neurons (nodes) joined with directed weighted edges which recently became popular as pattern classifiers \cite{lipton_critical_2015, graves_supervised_2012}. The network is usually activated by feeding an input that then spreads throughout the network along the edges. Many types of neural networks have evolved since its first appearance; however, they will fall under two main categories, the networks whose connections form cycles and the ones that are acyclic \cite{graves_supervised_2012}. RNNs are the type of neural networks that introduces the notion of time by using cyclic edges between adjacent steps. RNNs have been proposed in many forms including Elman networks, Jordan networks, and echo state networks \cite{elman_finding_1990, jordan_attractor_1990, jaeger_echo_2001, khalifa_sparse_2017}.
	
	\begin{figure}[!ht]
		\centering
		\includegraphics[width=\columnwidth]{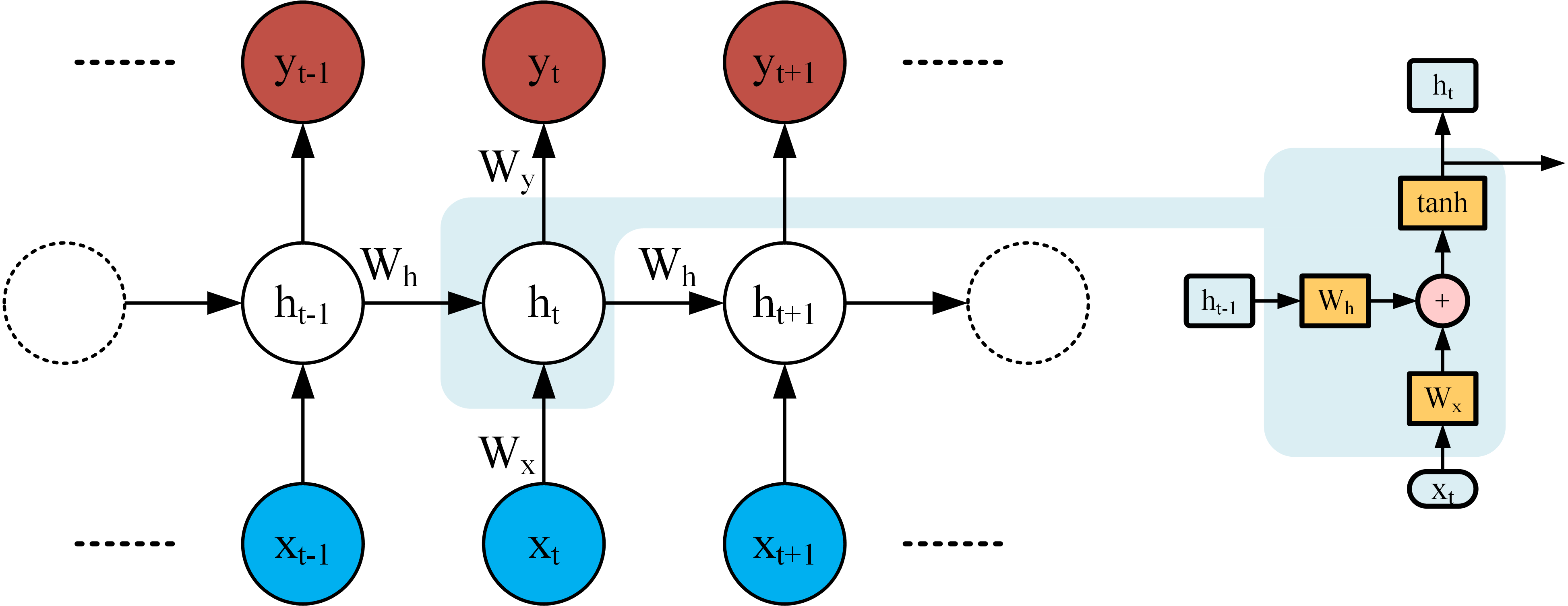}\\
		\caption{A simple RNN with a single hidden layer. At each time step $t$, output is produced through passing activations as in a feedforward network. Activations are passed to next node at time $t+1$ as well to achieve recurrence.}
		\label{fig:RNNex}
	\end{figure}
	
	As shown in Fig. \ref{fig:RNNex}, the hidden units at time t receive input from the current input $x_t$ and the previous hidden unit value $h_{t-1}$. The output $y_t$ is calculated using the current hidden unit value $h_t$. Time dependency is created between time steps by means of recurrent connections between hidden units. In a forward pass, all the computations are specified using the following two equations: $h_t= \sigma_h \left( W_x x_t + W_h h_{t-1} +b_h\right)$, $y_t= \sigma_y \left( W_y h_t + b_y\right)$; where $W_x$ and $W_y$ represent the matrices of weights between the hidden units and both input and output respectively and $W_h$ is the matrix of weights between adjacent time steps. $b_h$ and $b_y$ are bias vectors which allow offset learning at each node. Nonlinearity is introduced through the activation functions $\sigma_h$ and $\sigma_y$ which can be hyperbolic tangent function (tanh), sigmoid, or rectified linear unit (ReLU). In a simple RNN unit, tanh is usually used.
	
	\begin{figure}[h]
		\centering
		\begin{tabular}{c}
			\includegraphics[width=0.9\columnwidth]{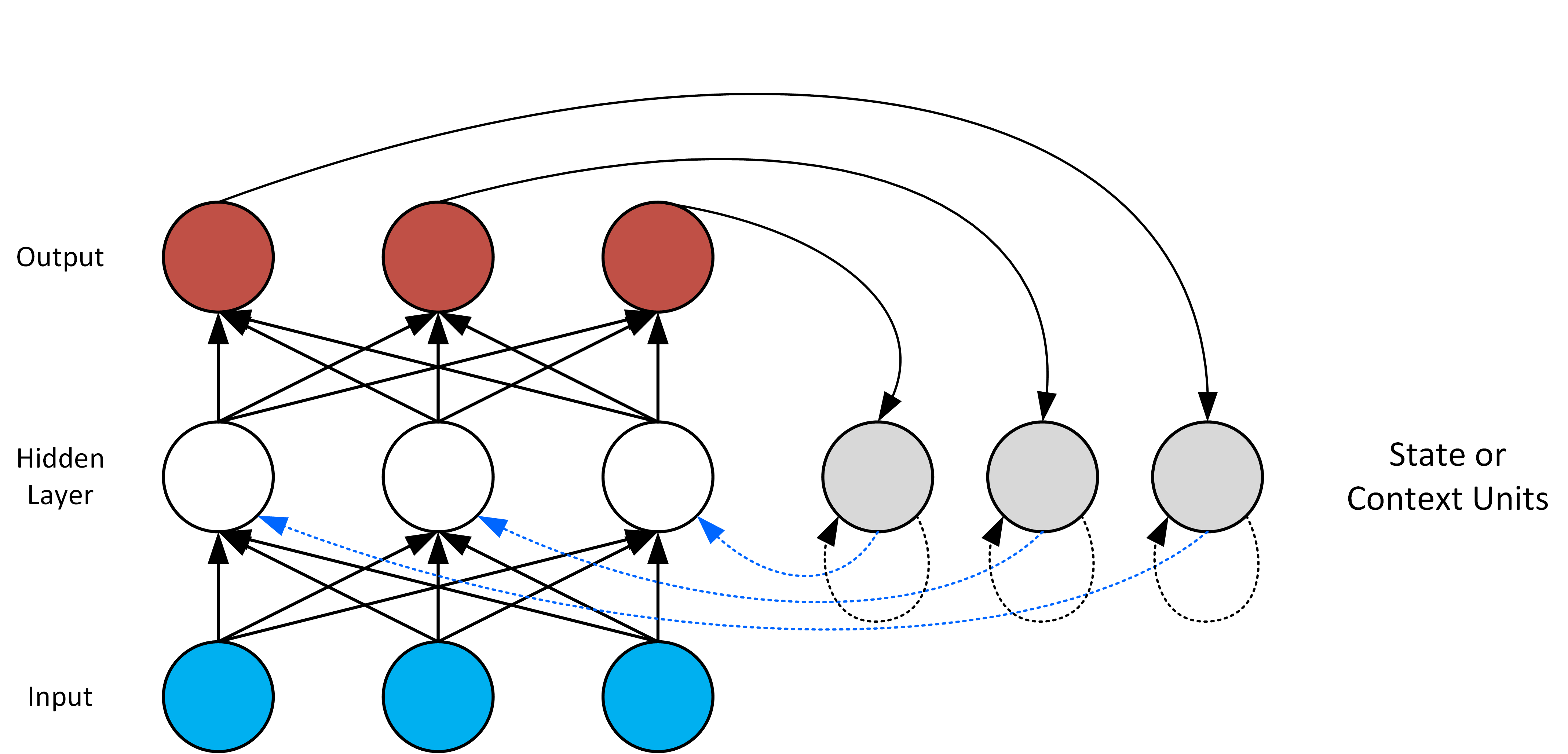}\\
			(a)\\
			\includegraphics[width=0.9\columnwidth]{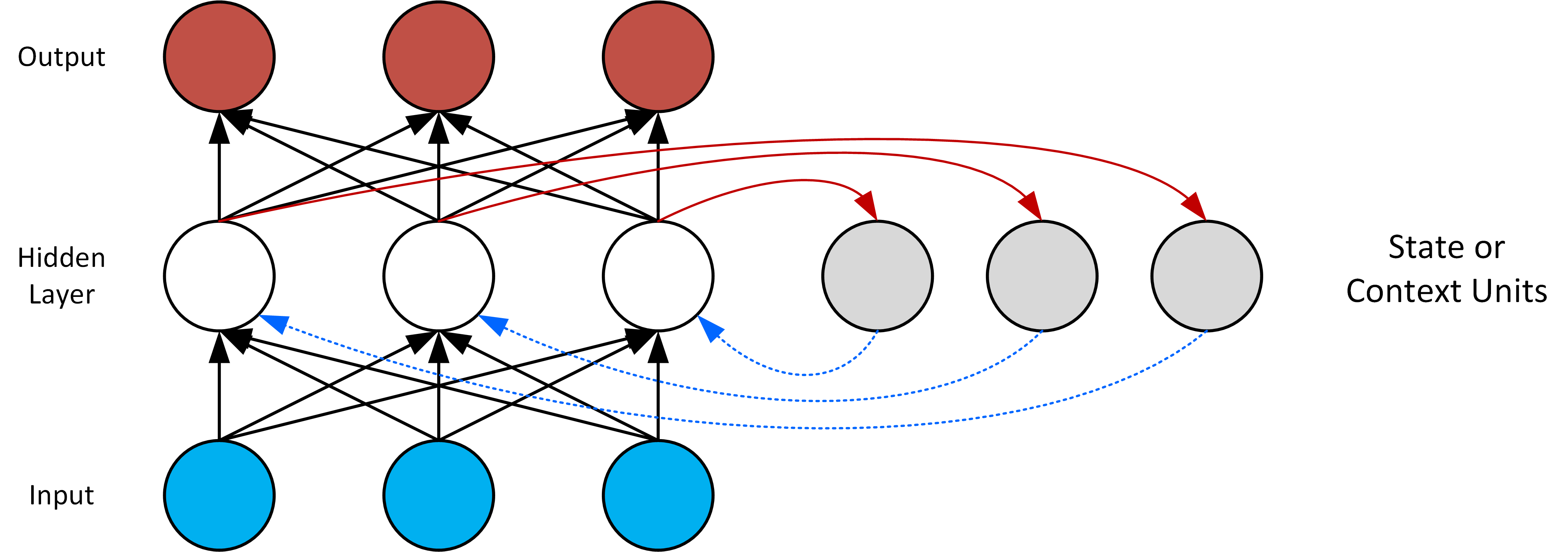}\\
			(b)
		\end{tabular}
		\caption{Early designs of RNNs. The dotted arrows represent the edges feeding at the next time step. (a) Jordan network. Output units are connected to context units that provide feedback at next time step to hidden units and themselves. (b) Elman network. Hidden units are connected to the context units that provide feedback to the hidden units only at the next time step.}
		\label{fig:earlyRNNs}
	\end{figure}

	\subsection{Early RNN Architectures}
		\citet{jordan_serial_1997} introduced an early form of recurrence in networks by adding extra "special" units called context or state units that feed values to the hidden units in the following time step. The network was as simple as a multi-layer feed-forward network with the context units taking input from the network output at the current time step and feed them back to themselves and the hidden units at the next time step as shown in Fig. \ref{fig:earlyRNNs} (a). The context units allow the network to remember its outputs at previous time steps and being self connected enables sending information across time steps without intermediate output perturbation \cite{lipton_critical_2015}. \citet{elman_finding_1990} also introduced a simple architecture in which the context units are associated with each each hidden layer unit at the current time step and give feedback to the same hidden unit at the next time step as shown in Fig. \ref{fig:earlyRNNs} (b). This notation of self-connected hidden units became the basis for the work and design of long-short term memory (LSTM) units \cite{hochreiter_long_1997}. This type of recurrence has been demonstrated to learn time dependencies by Elman \cite{elman_finding_1990}.
		
	\subsection{Training of RNNs}
\begin{figure}[!h]
			\centering
			\includegraphics[width=0.9\columnwidth]{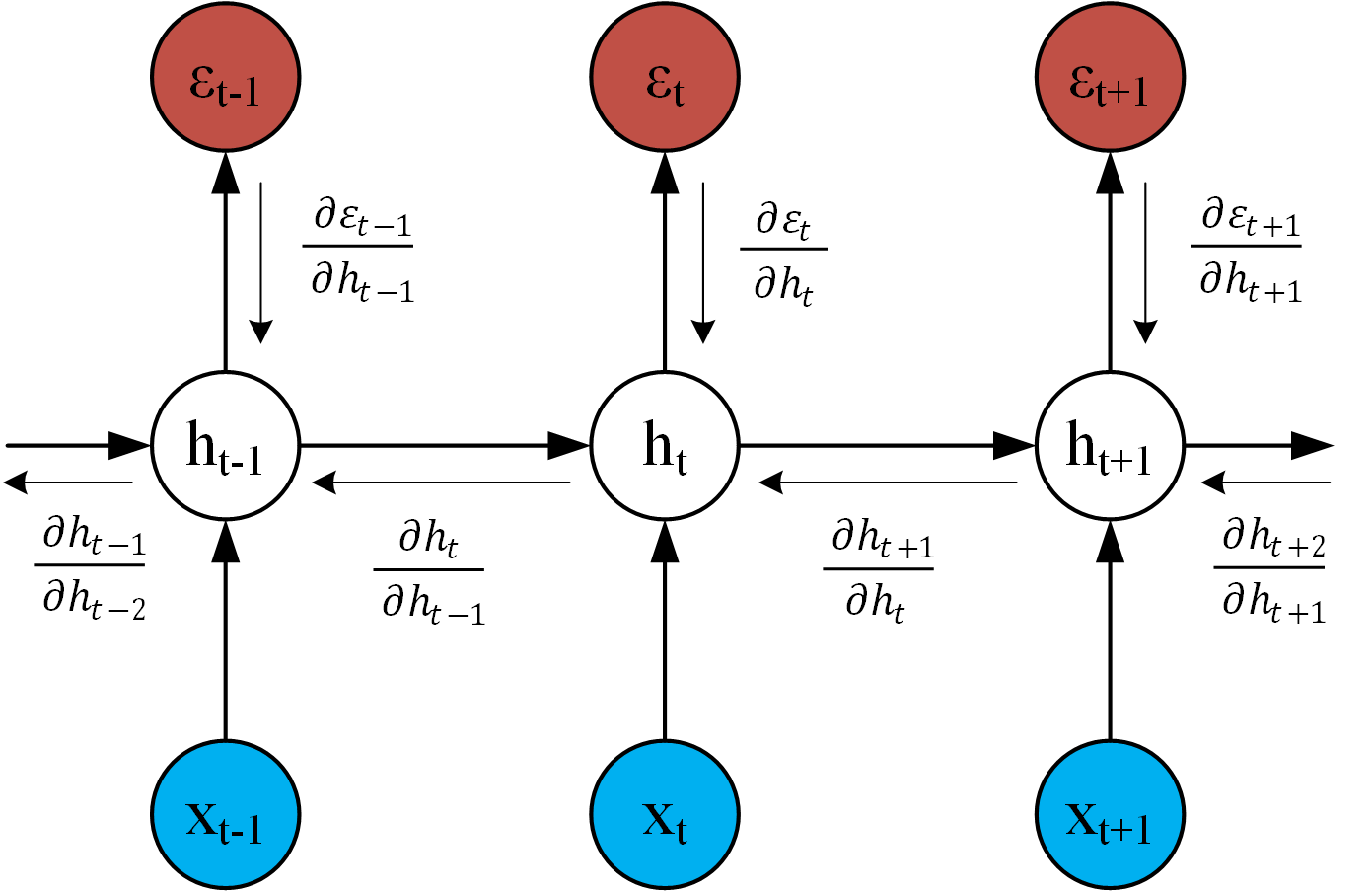}
			\caption{Unfolded recurrent neural network in time \cite{razvan_pascanu_difficulty_2013}. $\varepsilon_t$ denotes the error calculated from the output, $h_t$ represents the hidden state, and $x_t$ represents the input at time $t$.}
			\label{fig:RNNBPTT}
		\end{figure}

		The expression of  a generic RNN can be represented as $h_t = \mathcal{F}(h_{t-1}, x_t, \theta) = W_h\sigma_h(h_{t-1})+W_xx_t+b_h$\footnote{This formulation doesn't contradict with the previously mentioned formulation $\left( h_t= \sigma_h \left( W_x x_t + W_h h_{t-1} +b_h\right)\right) $ and both have the same behavior \cite{razvan_pascanu_difficulty_2013}.}, where $\theta$ refers to the network parameters $W_h$: recurrent weight matrix, $W_x$: input weight matrix, and $b_h$: the bias. Initial state $h_0$, is usually set to zero, provided by user, or learned. Network performance on a certain task is measured through a cost function $\varepsilon = \sum_{1\leq t\leq T}^{}\varepsilon_t$, where $\varepsilon_t = \mathcal{L}(h_t)$, $T$ is the sequence length (total number of time steps), and $\mathcal{L}$ is the cost operator that measures the performance of the network (e.g. squared error and entropy). Necessary gradients for optimization can be computed using backpropagation through time (BPTT), where the network is unrolled in time so that the application of backpropagation is feasible as shown in Fig. \ref{fig:RNNBPTT}.
		
		A gradient component $\frac{\partial\varepsilon}{\partial\theta}$ is calculated through the summation of temporal components as follows:
		\begin{equation*}
		\frac{\partial\varepsilon}{\partial\theta} = \sum\limits_{1\leq t\leq T}\frac{\partial\varepsilon_t}{\partial\theta}
		\end{equation*}
		\begin{equation*}
		\frac{\partial\varepsilon_t}{\partial\theta} = \sum\limits_{1\leq k\leq t} \left( \frac{\partial\varepsilon_t}{\partial h_t}\times\frac{\partial h_t}{\partial h_k}\times\frac{\partial h_k}{\partial\theta}\right) 
		\end{equation*}
		\begin{equation}\label{eqn:gradProd}
		\frac{\partial h_t}{\partial h_k} = \prod\limits_{t\geq i > k}\frac{\partial h_i}{\partial h_{i-1}} = \prod\limits_{t\geq i > k} W_h^T diag\left( \sigma_h'\left( h_{i-1}\right) \right)
		\end{equation}
		
		The effect that the network parameters ($\theta$) at step $k$ have over the cost at subsequent steps ($t>k$), can be measured through the temporal gradient component $\frac{\partial\varepsilon_t}{\partial h_t}\times\frac{\partial h_t}{\partial h_k}\times\frac{\partial h_k}{\partial\theta}$. In Eq. \ref{eqn:gradProd}, the matrix factors are in the form of a product of $t-k$ Jacobian matrices which will either explode or shrink to zero depending on whether the recurrent weights are greater or smaller than one \cite{razvan_pascanu_difficulty_2013}. The vanishing gradient is common when using sigmoid activations, while the exploding gradient is more common when using rectified linear unit activations \cite{razvan_pascanu_difficulty_2013,lipton_critical_2015}. Enforcing the weights through regularization to values that help avoid gradient vanishing and exploding, is one of the solutions to such a problem. Truncated backpropagation through time (TBPTT) is also used as another solution for exploding gradient through setting a maximum number of time steps through which the error is propagated \cite{lipton_critical_2015}.
		
	\subsection{Current RNN Designs}
		
\begin{figure}[!ht]
			\centering
			\begin{tabular}{c}
				\includegraphics[width=0.9\columnwidth]{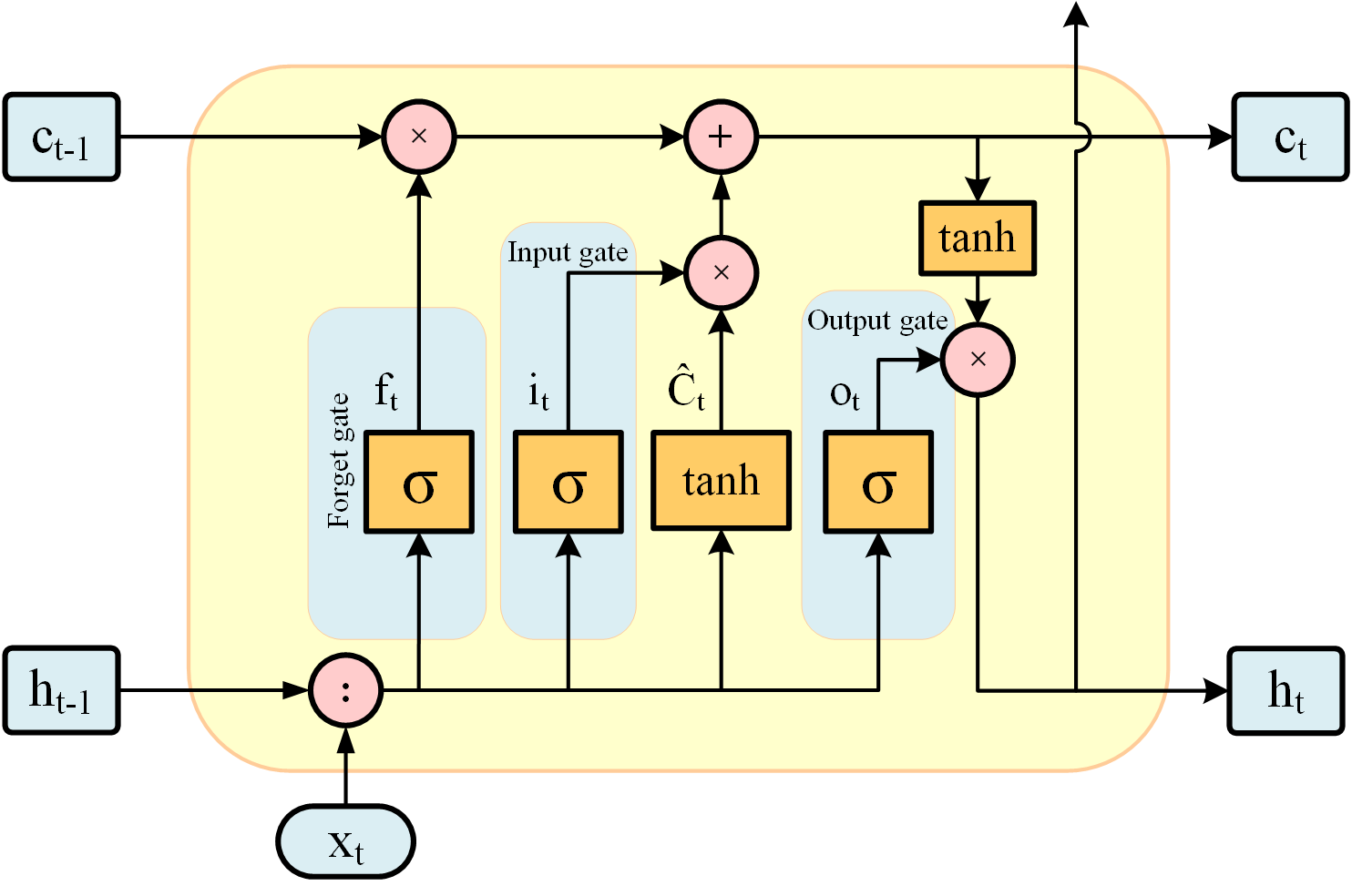}\\
				(a)\\
				\includegraphics[width=0.9\columnwidth]{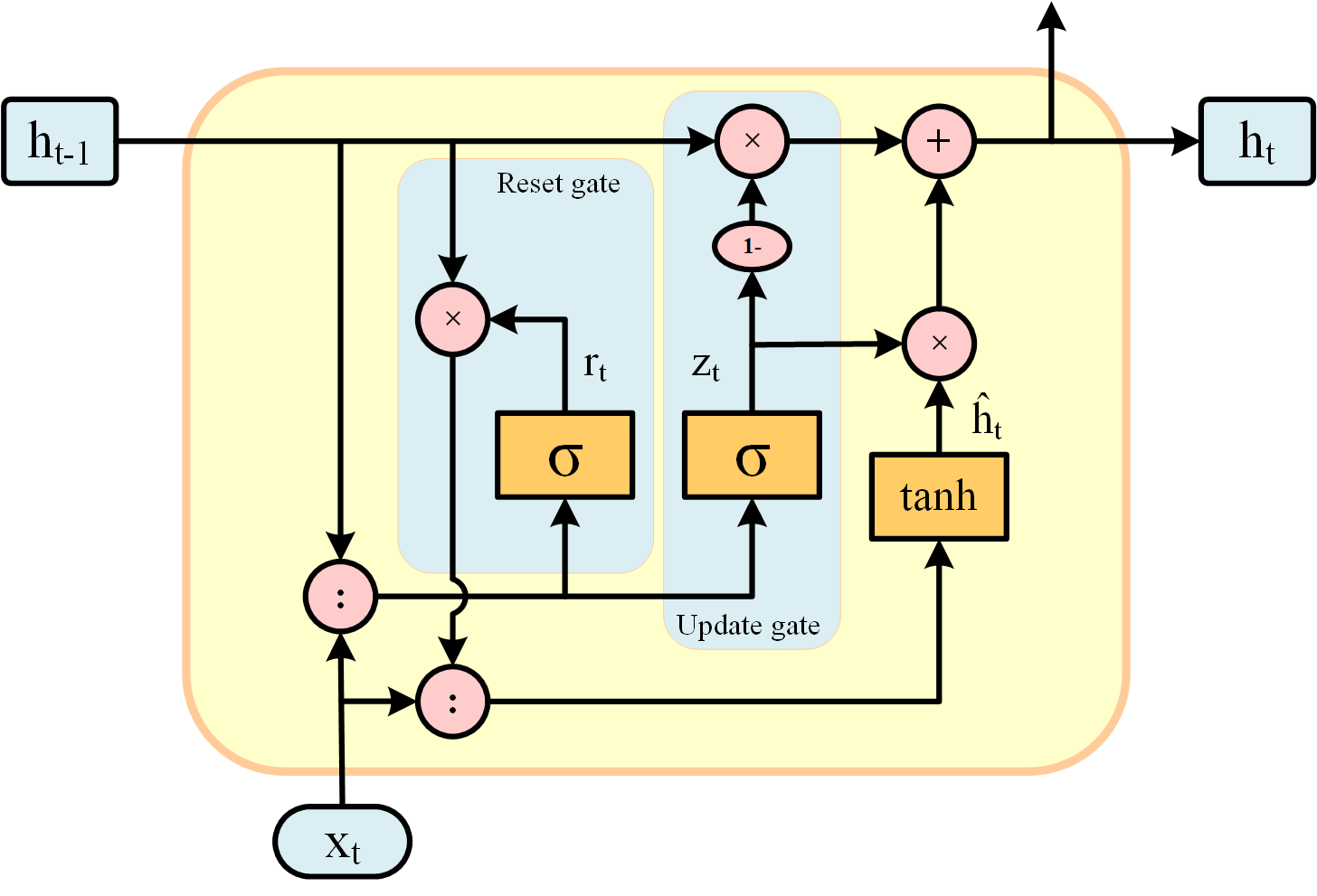}\\
				(b)
			\end{tabular}
			\caption{Current designs of RNNs. The symbols used in both diagrams are as follows, $:$ represents concatenation, $+$ represents element-wise summation, $\times$ represents element-wise multiplication, $\sigma$ represents a sigmoid activation, and $tanh$ represents a hyperbolic tangent (tanh) activation. (a) Schematic of an LSTM unit which is typically composed of three main parts, input, output, and forget gates. (b) Schematic of a GRU unit which is a simplified version of LSTM with only reset and update gates.}
			\label{fig:currRNNs}
		\end{figure}

		Although early designs of RNNs helped to map input into output sequences through using contextual information, this contextual mapping had limited range and the influence of input on hidden layers and thus output, either vanishes or blows up due to cycling through the network recurrent connections as described previously \cite{bengio_learning_1994}. Gradient vanishing/exploding problem has led to the emergence of new network designs that improve convergence \cite{graves_novel_2009,glorot_understanding_2010}. Of these designs, LSTM, gated recurrent units (GRU), and bidirectional RNNs (BRNN) have proved superiority in long-range contextual mappings and employing both future and past contexts to determine the output of the network \cite{lipton_critical_2015}. Both LSTM and GRU resemble a standard RNN but with each hidden node replaced by a complete cell as shown in Fig. \ref{fig:currRNNs}. They also employ a unity-weighted recurrent edge to ensure the transfer of gradient across time steps without decaying or exploding. LSTM forms the long-term memory through the weights which change slowly during training. On the other hand, short term memory is formed by transient activations that pass between successive node \cite{lipton_critical_2015}. GRU is an LSTM alternative that has a simpler structure and is faster to train; however, it still provides comparable performance to LSTM \cite{cho_learning_2014}. 
		
		In an LSTM unit, a forget gate is an adaptive gate whose output is squashed through a sigmoid activation in order to reset the memory blocks once they are out of date and prevent information storage for arbitrary time lags \cite{gers_learning_1999}. The input gate is a sigmoid activated gate whose function is to regulate the new information to be written to the cell state. The output gate is also a sigmoid activated gate that regulates the internal state after being dynamically customized through a $tanh$ activation to be forwarded as the unit output. In the same way, the GRU unit has a similar design; however, it doesn't have an output gate. It has a reset gate that works as a forget gate and an update gate to regulate the write operation into the unit output from both the state of the past time step and the input from the current time step.
		
		On the other hand, BRNNs resemble a standard RNN architecture as well but with two hidden layers instead of one and each hidden layer is connected to both input and output . One hidden layer passes activations in the forward directions (from the past time steps) and the other layer passes the activations in the backward direction (from future time steps). BRNN is in fact a wiring method for RNN hidden layers regardless of the type of the nodes, which makes it compatible with most RNN architectures including LSTM and GRU \cite{lipton_critical_2015,graves_novel_2009}.
		
\section{Critical Differences between HMMs and RNNs}
	As demonstrated in the previous sections, construction of hidden Markov models relies on a representing state space from which states are drawn. Scaling such system has long been considered to be difficult or infeasible even with the presence of dynamic programming solutions such as the Viterbi algorithm due to the quadratic complexity nature of the inference problem and transition probability matrix which causes the model parameter estimation and inference to scale in time as the size of the state space grows \cite{a_viterbi_error_1967}. Modeling long range dependencies also is impractical in HMMs as transitions occur from a state to the following with no memory of the previous state unless a new space is created with all possible cross-transitions at each time window which leads to exponential growth of the state space size \cite{lipton_critical_2015,graves_neural_2014}. On the other hand, the number of states that can be represented by a hidden layer in RNNs increases exponentially with the number of nodes in the layer leading to nodes that can carry information from contexts of arbitrary lengths. Moreover, despite of the exponential growth of the expressive power of the network, training and inference complexities only grow quadratically at most \cite{lipton_critical_2015}. From a theoretical point of view, RNNs can be efficient in the perception of long contexts; however, this comes at the cost of error propagation. Highly sampled inputs as in the case of raw waveforms, can lead to elongation of the range through which the error signal propagates, thus making the network hard to optimize and reducing the efficiency of computational acceleration tools such as GPUs \cite{schwab_beat_2017,stollenga_parallel_2015}.
	
\section{Event Detection in Electrocardiography}
	
		\begin{table*}[width=\linewidth,cols=4,pos=!ht]
		\caption{Summary of event detection work done in ECG event detection}
		\label{tab:ECGdetails}
		\begin{tiny}
			\begin{tabular}{p{1in}p{1.3in}p{2.5in}p{1.4in}}
				\toprule
				Publication & Event under investigation & Implementation details&Dataset\\
				\midrule
				\citet{gersch_pvc_1975}, \citeyear{gersch_pvc_1975} & Premature Ventricular Contraction (PVC) through R-R intervals & A three states Markov chain was used to model R-R interval (quantized as short, regular, or long) sequences and then the model is used to characterize rhythms through the probability that the observed R-R symbol sequence is generated by any of a set of models generated from multiple cardiac arrhythmias. Theb manuscript used a maximum likelihood approach to determine the arrhythmia type. & Clinical test data from patients with atrial fibrillation (AF)\\
				\hline
				\citet{coast_approach_1990}, \citeyear{coast_approach_1990} & Beat detection for arrhythmia analysis& A parallel combination of HMMs (one for each arrhythmia type), is used to classify arrhythmia. The classification process is inferred through determining the most likely path through the parallel models. All ECG waveform parts were included in the states of each model. The results reported in this study relied on single ECG channel and didn't include multi-channel ECG fusion. & The American Heart Association (AHA) ventricular arrhythmia database \cite{hermes_development_1980}\\
				\hline
				\citet{andreao_ecg_2006}, \citeyear{andreao_ecg_2006} & ECG beat detection and segmentation & An HMM was constructed for ECG beat with each waveform part represented in the model including the isoelectric parts (ISO, P, PQ, QRS, ST, T). Model parameters were estimated using Baum-Welch method and the number of states in each model were specified empirically to achieve a good complexity-performance compromise. The proposed segmentation in this study was based on a single channel but the authors provided insights about the possibility of adaptation with multi-channel fusion.& QT database \cite{laguna_database_1997}\\
				\hline
				\citet{sandberg_frequency_2008}, \citeyear{sandberg_frequency_2008}& Atrial fibrillation frequency tracking & An HMM is used for frequency tracking to overcome the corruption of residual ECG by muscular activity or insufficient beat cancellation. States of the HMM were used to represent the underlying frequencies in short-time Fourier transform while observations corresponded to the estimated frequency of specific time intervals from the signal. Experiments were performed on single channel simulated signals with inclusion of mutli-channel fusion. & Simulated atrial fibrillation signals with four different frequency trends: constant frequency, varying frequency, gradually decreasing frequency, and stepwise decreasing frequency.\\
				\hline
				\citet{oliveira_coupled_2017}, \citeyear{oliveira_coupled_2017}& Automatic segmentation (beat) of ECG and Phonocardiogram (PCG) & An ECG channel along with a phonocardiogram were fused in a single coupled HMM for beat detection. The coupled HMM was constructed to consider the high dynamics and non-stationarity of the signals where the channels were assumed to be co-dependent through past states and observations. Each of ECG and phonocardiogram was modeled using 4 states. This study introduced a decision-level fusion through combining two channels in a single HMM. The study also experimented two different coupled HMMs, a fully connected where transition can happen between any two states from both channels and a partially connected model where certain limitations were added over transitions through considering the prior knowledge of the relationship between heart sounds and ECG components. & A self-recorded dataset from healthy male adults.\\
				\hline
				\citet{ubeyli_combining_2009}, \citeyear{ubeyli_combining_2009}& Arrhythmia detection/classification & An Elman-based RNN is used for beat classification with the Levenberg-Marquardat algorithm for training (a least-squares estimation algorithm based on the maximum neighborhood idea). This model used power spectral density (calculated with three different methods; Pisarenko, MUSIC, and Minimum-Norm) of ECG signals as input.  All the models trained in this study, used feature-level fusion. & Four types of ECG beats obtained from Physiobank Database \cite{goldberger_physiobank_2000}.\\
				\hline
				\citet{zhang_patient-specific_2017}, \citeyear{zhang_patient-specific_2017}& Supraventriular and verntricular ectopic beat detection (SVEB and VEB) & An LSTM-based RNN preceded by a density-based clustering for training data selection from a large data pool. In this implementation, the authors fed the RNN with the current ECG beat and the T wave part from the former beat to automatically learn the underlying features. The RNN layers were followed by two fully connected layers in order to combine the temporal features and generate the desired output. This study only used a single channel ECG (limb lead II) with no multi-channel fusion. & MIT-BIH Arrhythmia database (MITDB) \cite{moody_impact_2001}.\\
				\hline
				\citet{xiong_robust_2017}, \citeyear{xiong_robust_2017}& Atrial fibrillation automatic detection& A 3 layer RNN was implemented to extract the temporal features from the raw ECG signals. No multi-channel fusion was performed in this study and only a single ECG channel was employed. &The 2017 PhysioNet/CinC Challenge dataset \cite{goldberger_physiobank_2000}.\\
				\hline
				\citet{schwab_beat_2017}, \citeyear{schwab_beat_2017}& Different cardiac arrhythmia classification/detection & In this work a combination of GRU and bidirectional LSTM (BLSTM) based RNNs and nonparameteric Hidden Semi-Markov Models (HSMM), was used for building the beat classification model and then a blender \cite{wolpert_stacked_1992} was used to combine the predictions from the models. No multi-channel fusion was performed in this study and only a single ECG lead was employed. &The 2017 PhysioNet/CinC Challenge dataset \cite{goldberger_physiobank_2000}.\\
				\hline
				\citet{zihlmann_convolutional_2017}, \citeyear{zihlmann_convolutional_2017}& Atrial fibrillation detection & A single layer LSTM-based convolutional RNN (CRNN) was constructed for atrial fibrillation detection in arbitrary length ECG recordings. This work employed the log spectrogram as an input to the CRNN to increase the accuracy. No multi-channel fusion was performed in this study and only a single ECG lead was used. & The 2017 PhysioNet/CinC Challenge dataset \cite{goldberger_physiobank_2000}.\\
				\hline
				\citet{limam_atrial_2017}, \citeyear{limam_atrial_2017}& Atrial fibrillation detection & A two layer LSTM-based CRNN was used for atrial fibrillation detection from single-lead ECG and heart rate. Feature-level fusion was performed after the convolutional neural network (CNN) layers to combine features from both inputs. The output from the RNN was used to either feed a dense layer to perform classification directly or train an SVM for classification and the results from both models were compared. & The 2017 PhysioNet/CinC Challenge dataset \cite{goldberger_physiobank_2000}.\\
				\hline
				\citet{chang_af_2018}, \citeyear{chang_af_2018}& Atrial fibrillation detection & A single layer LSTM-based RNN was constructed for atrial fibrillation detection in multi-lead ECG. This model also used spectrograms of the input ECG signals to feed the network. Feature-level fusion was performed to combine spectrograms of multi-lead ECG before feeding into the LSTM units. & Multiple datasets for atrial fibrillation and normal sinus rhythms \cite{petrutiu_abrupt_2007,moody_impact_2001,taddei_european_1992,nolle_crei-gard_1987,goldberger_physiobank_2000,bousseljot_nutzung_1995}.\\
				\hline
				\citet{lui_multiclass_2018}, \citeyear{lui_multiclass_2018}& Myocardial infarction classification & A deep single-layer LSTM based CRNN was used for classifying ECG beats from single-lead ECG. Multiple models were performed including a direct 4-class beat classifier from the LSTM CRNN via dense layers and 4-class beat classifier via the fusion of multiple one-versus-one binary classification networks using stacking. & The Physikalisch-Technische Bundesanstalt (PTB) diagnostic ECG database \cite{bousseljot_nutzung_1995} and the AF classification from a short single lead ECG recording: Physionet/computing in cardiology challenge 2017 database (AF-Challenge) \cite{clifford_af_2017}. \\
				\hline
				\citet{singh_classification_2018}, \citeyear{singh_classification_2018}& Arrhythmia detection & 3 models were built for arrhythmia detection, each of them is based on a different type of RNN. Regular RNNs, GRU, and LSTM were used for each of the three models. Each model included 3 layers of different unit sizes with a dense layer to generate a classification output (normal/abnormal). No multi-channel fusion was performed in this study and only a single ECG lead (ML2) was employed. & MIT-BIH Arrhythmia database (MITDB) \cite{moody_impact_2001}.\\
				\bottomrule
			\end{tabular}
		\end{tiny}
	\end{table*}
		
	ECG is the graphical interpretation of skin-recorded electrical activity of the electric field originating in the heart \cite{kadish_accaha_2001}. ECG provides information that is not readily available through other methods about heart activity and is considered the most commonly used procedure in the diagnosis of cardiac diseases due to the fact that it is non-invasive, simple, and cost-effective. This makes ECG subject to intense research related to the automatic analysis to reduce the subjectivity and the time spent on interpreting hours of recordings \cite{h_crawford_accaha_1999,andreao_ecg_2006, sayed_arrhythmia_2015}. ECG is a time periodic signal, which allows to mark out an elementary beat that constitutes the basis for ECG signal analysis \cite{andreao_ecg_2006}. For instance, heart rate can be estimated through the detection of QRS-complex from an ECG signal and the time interval between successive QRS-complexes (also known as R-R interval) can be used to detect premature ectopic beats \cite{kadish_accaha_2001}. In that sense, ECG beat detection is considered fundamental for most of the automated analysis algorithms. A detailed description of the recent publications that cover event detection in ECG using different methods, is included in Table \ref{tab:ECGdetails}.
	
\section{Event Detection in Electroencephalography}
	EEG is mostly a non-invasive technique to measure the electrical activity of the brain through a set of electrodes placed on the subject's scalp. EEG exhibits highly non-stationary behavior and significant non-linear dynamics \cite{schomer_niedermeyers_2012}. The excitatory and inhibitory postsynaptic potentials of the cortical nerve cells are considered the main source of EEG signals \cite{subha_eeg_2010}. EEG can be invasive if acquired using subdural electrode grids or using depth electrodes and is called intracranial EEG (iEEG); however, typical EEG signals are recorded from scalp locations specified by the 10-20 electrode placement criterion designed by the International Federation of Societies for Electroencephalography and have an amplitude of 10-100 $\mu V$ and a frequency range of 1-100 Hz \cite{schomer_niedermeyers_2012, subha_eeg_2010}. EEG signals are used in the diagnosis of multiple neurological disorders including epilepsy, lesions, tumors, and depression and their characteristics depend strongly on the age and state of the subject. There are multiple events that influence EEG and require the tedious job of analyzing hours of recordings to be extracted. These events range from the diagnosis/detection of certain seizures and syndromes to the tasks of brain computer interface (BCI). These events include the different sleep stages and sleep disorders, epileptic seizures, the effect of music or other artifacts, and the motor imagery tasks.
	
	\subsection{Sleep Staging in EEG}
		
		\begin{table*}[width=\linewidth,cols=4,pos=!ht]
			\centering
			\caption{Summary of EEG-based sleep staging.}
			\label{tab:EEGSleep}
			\begin{tiny}
				\begin{tabular}{p{0.9in}p{1.1in}p{3.2in}p{1.0in}}
					\toprule
					Publication & Event under investigation & Implementation details&Dataset\\
					\midrule
					\citet{flexerand_automatic_2002}, \citeyear{flexerand_automatic_2002} & Sleep staging in combined EEG and EMG & A three state (wakefulness, deep sleep, and rapid eye movement sleep) Gaussian observation HMM (GOHMM) was used and sleep stages were represented as mixtures of the basic three states. The probability of being in any of the three states was computed for 1 sec windows so that a continuous probability monitoring can be achieved. Expectation-maximization algorithm was used for parameter estimation and the Viterbi algorithm was used to calculate the posteriori estimate for being in each state. Feature-level fusion was performed on features from EEG channels (C3 and C4) and EMG. &  Nine whole-night sleep recordings from a group of nine healthy adults.\\
					\hline
					\citet{flexer_reliable_2005}, \citeyear{flexer_reliable_2005} & Sleep staging in single channel EEG (C3) & A three state (wakefulness, deep sleep, and rapid eye movement sleep) Gaussian observation HMM (GOHMM) was used and sleep stages were represented as mixtures of the basic three states. The probability of being in any of the three states was computed for 1 sec windows so that a continuous probability monitoring can be achieved. Expectation-maximization algorithm was used for parameter estimation and the Viterbi algorithm was used to calculate the posteriori estimate for being in each state. No multi-channel fusion was performed in this study and only a single EEG channel was used. & Two datasets were used, the first consists of 40 whole night sleep recordings from healthy adults and the second consists of 28 whole night sleep recordings of healthy adults.\\
					\hline
					\citet{doroshenkov_classification_2007}, \citeyear{doroshenkov_classification_2007} & Sleep staging using two channel EEG (Fpz-Cz and Pz-Oz) & A six state HMM was constructed for the purpose of sleep staging. Baum-welch algorithm was used for model's parameter estimation and the Viterby algorithm for state sequence decoding. Feature-level fusion was performed for features calculated from the two EEG channels. & Sleep-EDF database \cite{kemp_analysis_2000}.\\
					\hline
					\citet{bianchi_probabilistic_2012}, \citeyear{bianchi_probabilistic_2012} & Sleep cycle (quantifying probabilistic transitions between stages and multi-exponential dynamics) and fragmentation in case of apnea in PSG & An eight state HMM was constructed for sleep-wake activity. The connectivity between states was inferred through exponential fitting of subsets of the pooled bouts and adjacent-stage analysis. & Sleep Heart Health Study database \cite{quan_sleep_1997}.\\
					\hline
					\citet{pan_transition-constrained_2012}, \citeyear{pan_transition-constrained_2012} & Sleep staging using central EEG (C3-A2), chin electromyography (EMG), and electrooculogram (EOG) & A six state transition-constrained discrete HMM was constructed for sleep staging. Thirteen features were utilized including temporal and spectrum analyses of the EEG, EOG and EMG signals with feature-level fusion employed. & PSG including six channel EEG, EOG, EMG, and ECG signals, was obtained from 20 healthy subjects.\\
					\hline
					\citet{yaghouby_quasi-supervised_2015}, \citeyear{yaghouby_quasi-supervised_2015} & Sleep staging and scoring (quasi-supervised) in PSG & A five state Gaussian HMM was constructed for sleep staging with Baum-Welch algorithm for parameter estimation. In this implementation, feature-level fusion was achieved through feeding augmented vector of PSG features and human rated scores into the estimation algorithm in order to obtain the parameters to maximize the likelihood that a model with larger number of states explains the data. & Sleep-EDF database \cite{kemp_analysis_2000}.\\
					\hline
					\citet{onton_visualization_2016}, \citeyear{onton_visualization_2016} & Sleep staging in 2-channel home EEG (FP1-A2 and FP2-A2) and electrodermal activity (EDA) & A five state Gaussian HMM was constructed for sleep staging with expectation-maximization algorithm for parameter estimation and the Viterbi algorithm to find the maximum a posteriori estimate of state sequence. In this implementation, the relative power across the entire night was averaged in five frequency bands and fed into the model (feature-level fusion). & A self recorded data from 51 participants who were medication-free and self-reported asymptomatic sleepers and wit no history of neurologic or psychiatric disorders.\\
					\hline
					\citet{davidson_detecting_2005}, \citeyear{davidson_detecting_2005} & Behavioral microsleep detection in EEG (P3-01 and P4-02) & This study utilized an LSTM-based RNN to detect the lapses in visuomotor performance associated with behavioral microsleep events. The network used the power spectral density of 1 sec windows of the used two channels (calculated using the covariance method) with feature-level fusion in place to combine data. The network included 6 LSTM blocks of 3 memory cells each. & A self-recorded dataset from 15 subjects performing visuomotor tracking task. \\
					\hline
					\citet{hsu_automatic_2013}, \citeyear{hsu_automatic_2013} & Automatic deep sleep staging in single channel EEG (Fpz-Cz) & This study utilized an Elman recurrent neural network that works on the energy features extracted from a single channel EEG to perform 5-level sleep staging. No multi-channel fusion was employed in this study. &Sleep-EDF database \cite{kemp_analysis_2000}. \\
					\hline
					\citet{supratak_deepsleepnet_2017}, \citeyear{supratak_deepsleepnet_2017} & Automatic sleep staging in single channel EEG (Fpz-Cz or Pz-Oz) & A convolutional RNN (CRNN) was constructed to work directly of the raw signal data. Two branches of CNN, each of 4 layers, were used for representation learning and their outputs were combined and fed into a two layer LSTM-based BRNN with skip branch to generate the sleep stage. No multi-channel fusion was employed in this study. & Montreal Archive of Sleep Studies (MASS) \cite{oreilly_montreal_2014} and Sleep-EDF database \cite{kemp_analysis_2000}. \\
					\hline
					\citet{biswal_sleepnet_2017}, \citeyear{biswal_sleepnet_2017} & Automatic sleep staging & Raw EEG signals were split into 30-seconds windows, then the spectrogram and expert defined features were extracted and fused at the feature-level. The best accuracy reported among different RNN architectures, was reported for a 5-layer LSTM-based RNN. This study presented also an LSTM-based CRNN architecture to extract spatial features automatically and then pass them to the RNN part for temporal context extraction. & 10,000 PSG studies with multi-channel EEG data (F3, F4, C3, C4, O1 and O2 referenced to the contralateral mastoid, M1 or M2). \\
					\hline
					\citet{phan_automatic_2018}, \citeyear{phan_automatic_2018} & Automatic deep sleep staging in single channel EEG (Fpz-Cz) & A two-layer GRU-based BRNN was constructed to learn temporal features from the single channel EEG. This implementation included an attention mechanism that was applied on the BRNN output features. The weighted output was then used to feed a linear SVM classifier. No multi-channel fusion has been employed in this study. &Sleep-EDF database \cite{kemp_analysis_2000}. \\
					\hline
					\citet{bresch_recurrent_2018}, \citeyear{bresch_recurrent_2018} & Sleep staging in single-channel EEG & An LSTM-based CRNN with 3 CNN layers and 3 LSTM layers, was built to process 30-seconds windows of raw EEG data (FPz, left EOG, and right EOG referenced to M2). No multi-channel fusion has been employed in this study. & The SIESTA database \cite{klosh_siesta_2001} and a self-recorded dataset with 147 recordings from 29 healthy subjects.\\
					\hline
					\citet{phan_seqsleepnet_2019}, \citeyear{phan_seqsleepnet_2019} & Automatic sleep staging & This study featured multi-modality fusion on the feature level between EEG, EOG, and EMG. All were split into windows and converted into time-frequency representation using filter banks. The fused data were fed into a BRNN that is used to encode the features, then the output is passed through an attention layer followed by another BRNN that performs the cclassification of the sleep stage. & Montreal Archive of Sleep Studies (MASS) Dataset \cite{oreilly_montreal_2014}. \\
					\hline
					\citet{michielli_cascaded_2019}, \citeyear{michielli_cascaded_2019} & Automatic sleep staging in single channel EEG & A dual branch LSTM-based RNN was constructed for the classification of 5 different sleep stages. the network starts with a preprocessing and feature extraction stages and then the data is distributed over two branches. The first branch uses mRMR for feature selection followed by a one layer LSTM and fully connected layer to classify between 4 classes only (W, N1-REM, N2 and N3). The second branch uses PCA for feature selection followed by a 2 layer LSTM and a fully connected layer for binary classification. The LSTM in the second branch takes the classification output from the first branch to consider only the combined stage N1-REM for separation. No multi-channel fusion has been employed in this study. & Sleep-EDF database \cite{goldberger_physiobank_2000}. \\
					\hline
					\citet{sun_two-stage_2019}, \citeyear{sun_two-stage_2019} & Sleep staging in single channel EEG & A two stage network was built to perform the classification. The first stage is time distributed stage that included two parallel branches, the first included a window deep belief network for feature extraction followed by a dense layer and a second branch with hand-crafted features extraction then a dense layer. The two branches were then fused through another dense layer and fed as an input to an LSTM-based BRNN (the second stage) to generate the classes. & Sleep-EDF database \cite{goldberger_physiobank_2000}.\\
					\bottomrule
				\end{tabular}
			\end{tiny}
		\end{table*}
		
		Sleep is an essential part of the human life cycle and plays a vital role in maintaining most of the body functionality \cite{sheldon_principles_2005}. Sleep disorders include problems with initiating sleep, insomnia, and sleep apnea syndrome (SAS) \cite{kang_state_2018}. Diagnosis of sleep disorders can be done through identifying sleep stages in an overnight polysomnogram (PSG) which utilizes EEG as one of its sensing modalities \cite{roebuck_review_2013}. Visual scoring of the PSG components is the basic way to categorize sleep epochs and as any manual rating, it suffers from subjectivity and inter-rater tolerance. Many attempts have been proposed in the literature to remedy the problems of expert-based visual scoring of the different components of PSG. The attempts employed multiple algorithms to achieve automatic sleep staging including Markov models and neural networks. Here, we list the recent publications (Table \ref{tab:EEGSleep}) for sleep staging and the detailed description of the methods used within the scope of our review.
	
	\subsection{Epilepsy detection in EEG} \label{ssec:EpEEG}
		Epilepsy is one of the episodic disorders of the brain that is characterized by recurrent seizures, unjustified by any known immediate cause \cite{commission_on_epidemiology_and_prognosis_guidelines_1993,lytton_computer_2008}. Epileptic seizure is the clinical manifestation that results from the abnormal excessive discharge of some set of neurons in the brain \cite{commission_on_epidemiology_and_prognosis_guidelines_1993}. The seizure consists of transient abnormal alterations of sensory, motor, consciousness, or psychic behavior \cite{commission_on_epidemiology_and_prognosis_guidelines_1993,lytton_computer_2008}. Around 80\% of the epileptic seizures can be effectively treated if early discovered \cite{abdullah_seizure_2012}. Although seizure activity can be easily distinguished in EEG as transient spikes and relatively quiescent periods, it is a time-consuming process and needs clinicians to devote a tremendous amount of time going through hours and days of EEG activity \cite{smart_semi-automated_2015}. An efficient and reliable seizure prediction/detection method can be of a great help for the diagnosis, treatment, and even early warning for patients to stop activities that might be of a significant danger during an episode like driving. Several methods have been proposed for seizure prediction, at which EEG signal features are temporally analyzed and compared to heuristic thresholds to trigger a warning for seizures; however, these methods lack generalization when investigated on extensive datasets \cite{santaniello_bayesian_2011,mormann_predictability_2005,maiwald_comparison_2004,mcsharry_prediction_2003,lai_controlled_2004,winterhalder_seizure_2003}. This can be referred to using feature sets that are not highly affected by the transition from seizure-free to peri-ictal or seizure states or simply the effect cannot be tracked using low-order statistics \cite{santaniello_bayesian_2011}. Therefore, stochastic-based models, multivariate analysis, and long-range analysis methods were investigated to provide better performance and generalization for EEG-based epileptic seizure prediction. In Table \ref{tab:EEGEpilepsy}, we review the recent publications that use HMMs and RNNs for seizure prediction.
		
		\begin{table*}[width=\linewidth,cols=4,pos=h]
			\centering
			\caption{Summary of EEG-based seizure prediction.}
			\label{tab:EEGEpilepsy}
				\begin{tiny}
					\begin{tabular}{p{1in}p{1.1in}p{2.8in}p{1.3in}}
						\toprule
						Publication & Event under investigation & Implementation details&Dataset\\
						\midrule
						\citet{wong_stochastic_2007}, \citeyear{wong_stochastic_2007} & Evaluation framework for seizure prediction in iEEG & A three state HMM (baseline, detected, and seizure) was constructed to evaluate the prediction algorithms of epileptic seizures. The prediction algorithm is used to generate a binary sequence which is combined with the ground truth (binary detector outputs plus gold-standard human seizure markings) and converted into a trinary observation sequence. The trinary vector is used to train the HMM using Baum-Welch which is then used to Viterbi decode the observation sequences into the hidden states sequence. A hypothesis test that a statistical association exists between the detected and seizure states, is performed through counting the transitions from detected state into seizure states in the HMM output. & iEEG data collected from patients diagnosed with mesial temporal lobe epilepsy using 20-36 surgically implanted electrodes on the brain or brain substance \cite{gardner_one-class_2006}.\\
						\hline
						\citet{santaniello_bayesian_2011}, \citeyear{santaniello_bayesian_2011} & Early detection of seizures in iEEG from a rat model & Multichannel iEEG were used and Welch's cross power spectral density was calculated over windows of 3 sec for each pair of channels which were used as input for the detection model. A two state HMM was constructed to map the iEEG signals into either normal or peri-ictal states. Baum-Wlech algorithm was used for parameter estimation and a Bayesian evolution model was used determine the time of state transition. & Data collected from male Sprague-Dawley rats with four implanted skull screw EEG electrodes placed bifrontally and posteriorly behind bregma and a fifth depth electrode placed in hippocampus, were collected and used for this study.\\
						\hline
						\citet{direito_modeling_2012}, \citeyear{direito_modeling_2012} & Identification of the different states of epileptic brain & The relative power in EEG sub-bands (delta, theta, alpha, beta, and gamma) was calculated and used for computing the topographic maps of each sub-band. The maps were then segmented and used overtime to train a 4 state (preictal, ictal, postictal and interictal) HMM. The Baum–Welch algorithm was used to train the model and the Viterbi algorithm to decode the state-sequence. & EPILEPSIAE database \cite{ihle_epilepsiae_2012}. \\
						\hline
						\citet{abdullah_seizure_2012}, \citeyear{abdullah_seizure_2012} & Seizure detection in iEEG & A three state discrete HMM was built to classify iEEG segments into one of three states (ictal, preictal, and interictal). Seven level decomposition stationary wavelet transform (SWT) was applied on the signals (as input features for the model) and a code book was created to perform vector quantization. Baum-Welch algorithm was used for model parameter estimation and the Viterbi algorithm for recognition. This study employed a feature-level fusion model to feed the data into the prediction model. & Freiburg Seizure Prediction EEG (FSPEEG) database \cite{maiwald_comparison_2004}.\\
						\hline
						\citet{smart_semi-automated_2015}, \citeyear{smart_semi-automated_2015} & Seizure detection in scalp EEG & This study used a 5 sec sliding window with 1 sec increments to process the EEG signals. A set of 45 measurements was calculated for each sliding window then principal component analysis (PCA) was used to reduce dimensionality. One of the used models was HMM, particularly a two state (seizure and non-seizure) HMM was constructed to perform the detection. Baum-Welch was used here as well to estimate the model parameters.This study used a feature-level fusion model for multi-channel EEG data to feed the data into the prediction model.  & CHB-MIT Scalp EEG Database \cite{shoeb_application_2009}. \\
						\hline
						\citet{petrosian_recurrent_2000}, \citeyear{petrosian_recurrent_2000} & Onset detection of epileptic seizures in both scalp and intracranial EEG & Both raw EEG data and their wavelet transform "daub4" were used in training an Elman RNN. This study used a feature-level fusion model for multi-channel EEG data to provide an input for the RNN. & Scalp and iEEG data were collected from two patients who were undergoing long-term electrophysiological monitoring for epilepsy. \\
						\hline
						\citet{guler_recurrent_2005}, \citeyear{guler_recurrent_2005} & Identification of subject condition in terms of epilepsy (healthy, epilepsy patient during seizure-free interval, and epilepsy patient during seizure episode) using surface and intracranial EEG & Lyapunov exponents of the EEG signals were used to train an Elman RNN for the identification task. This study used a feature-level fusion model for multi-channel EEG data to train the RNN. & Publicly available epilepsy dataset by University of Bonn \cite{andrzejak_indications_2001}.\\
						\hline
						\citet{kumar_automated_2008}, \citeyear{kumar_automated_2008} &  Automatic detection of epileptic seizure in surface and intracranial EEG & Wavelet and spectral entropy were extracted from the EEG signals and used to train an Elman RNN. This study used a feature-level fusion model for multi-channel EEG data to train the RNN. & Publicly available epilepsy dataset by University of Bonn \cite{andrzejak_indications_2001}.\\
						\hline
						\citet{minasyan_patient-specific_2010}, \citeyear{minasyan_patient-specific_2010} & Automatic detection of epileptic seizures prior to or immediately after clinical onset in scalp EEG & A set of time domain, spectral domain, wavelet domain, and information theoretic features were used to train an ELman RNN per each channel of the EEG and the output is combined in time and space through a decision making module that performs a decision-level fusion in order to declare a seizure event if N out of M channels declared it. & EEG dataset from 25 patients hospitalized for long-term EEG monitoring in five centers including Thomas Jefferson University, Dartmouth University, University of Virginia, UCLA and University of Michigan medical centers.\\
						\hline
						\citet{naderi_analysis_2010}, \citeyear{naderi_analysis_2010} & Automatic detection of epileptic seizure in surface and intracranial EEG & Power spectral density was calculated for EEG signals using Welch method then a dimensionality reduction algorithm was applied and the output was used to train an ELman RNN. This study used a feature-level fusion model for multi-channel EEG data to train the RNN. & Publicly available epilepsy dataset by University of Bonn \cite{andrzejak_indications_2001}.\\
						\hline
						\citet{vidyaratne_deep_2016}, \citeyear{vidyaratne_deep_2016} & Automated patient specific seizure detection using scalp EEG & The preprocessed (denoised) EEG signals were segmented into 1 sec non overlapping epochs and used to train a BRNN. Data from all channels were used simultaneously (feature-level fusion model). & CHB-MIT Scalp EEG Database \cite{shoeb_application_2009}.\\
						\hline
						\citet{talathi_deep_2017}, \citeyear{talathi_deep_2017} & Epileptic seizures detection & Single-channel EEG data (no multi-channel fusion) were used to train a GRU-based RNN that classifies each EEG segment into one of three states: healthy, inter-ictal, or ictal. Two layers of GRU were used, the first was followed by a fully connected layer and the second was followed by a logistic regression classification layer.  &Publicly available epilepsy dataset by University of Bonn \cite{andrzejak_indications_2001}.\\
						\hline
						\citet{Golmohammadi_gated_2017}, \citeyear{Golmohammadi_gated_2017} & Epileptic seizure detection & Linear frequency cepstral coefficient feature extraction was performed for the EEG data and used to feed a CRNN that is based on a bidirectional LSTM. Features from multi-channel EEG were fused prior to feeding into the CRNN. The network used in this study employed both 2D and 1D CNN at different stages. Another network where LSTM was replaced with GRU was devloped as well for comparison. & A subset of the TUH EEG Corpus (TUEEG) \cite{obeid_temple_2016} that has been manually annotated for seizure events \cite{Golmohammadi_TUH_2017}.\\
						\hline
						\citet{raghu_classification_2017}, \citeyear{raghu_classification_2017} & Epileptic seizures classification & This study developed two techniques that are based on Elman RNN that works on features extracted from EEG signals. The first technique used wavelet decomposition with the estimation of log energy and norm entropy to feed the RNN classifier (normal vs preictal). The second way extracted the log energy entropy to feed the RNN classifier. &Publicly available epilepsy dataset by University of Bonn \cite{andrzejak_indications_2001}. \\
						\hline
						\citet{abdelhameed_deep_2018}, \citeyear{abdelhameed_deep_2018} & Epileptic seizure detection & This study used raw EEG signals to feed a 1D CRNN that is based on bidirectional LSTM to classify EEG segments into one of two states (normal-ictal and normal-ictal-interictal).  &Publicly available epilepsy dataset by University of Bonn \cite{andrzejak_indications_2001}. \\
						\hline
						\citet{daoud_deep_2018}, \citeyear{daoud_deep_2018} & Epileptic seizure prediction & This study used raw EEG signals to feed a 2D CRNN that is based on a bidirectional LSTM to classify EEG segments into one of two classes (preictal and interictal). & A dataset recorded at Children's Hospital Boston which is publicly available \cite{shoeb_application_2009,goldberger_physiobank_2000}.\\
						\hline
						\citet{hussein_optimized_2019}, \citeyear{hussein_optimized_2019} & Epileptic seizures detection & This study developed an LSTM-RNN that takes raw EEG signals as input in order to create predictions. The network was composed of a one layer LSTM followed by a fully connected layer and an average pooling layer to combine the temporal features and then an output softmax layer.  &Publicly available epilepsy dataset by University of Bonn \cite{andrzejak_indications_2001}.\\
						\bottomrule
					\end{tabular}
				\end{tiny}
		\end{table*}
	
	\subsection{BCI Tasks in EEG}
		Motor imagery alters the the neural activity of the brain's sensorimotor cortex in a way that is as observable as if the movement was really executed \cite{pfurtscheller_motor_2001}. Identification of the transient patterns in EEG signals during the different motor imagery tasks like imagining the movement of one of the limbs, is recognized among the most promising and widely used techniques of BCI \cite{leuthardt_brain-computer_2004,schalk_brain-computer_2011,sayed_characterization_2017,sayed_extracting_2017}. This is referred to the relatively low cost of the systems used and the high temporal resolution \cite{sayed_characterization_2017}. This type of BCI is called asynchronous BCI because the subject is free to invoke specific thought \cite{pfurtscheller_motor_2001}. On the other hand, synchronous BCI includes the generation of specific mental states in response to external stimuli \cite{pfurtscheller_motor_2001}. EEG analysis for BCI applications includes the processing of EEG oscillatory activity and the different shifts in its sub-bands in addition to the event-related potentials like VEP and P300 \cite{pfurtscheller_motor_2001,donchin_mental_2000}. Many modeling schemes have been introduced to solve the of multi-class BCI problem; however, most of them process EEG signals in short windows where stationarity is assumed, which limits the modeling process and excludes the dynamic EEG patterns such as desynchronization \cite{pfurtscheller_motor_2001}. To overcome such a limitation, probabilistic models like HMMs and models capable of representing long range dependencies have been proposed into the implementation of BCI systems. As follows in Table \ref{tab:EEGBCI}, we list the recent work the relies on HMMs and RNNs in BCI systems and uses EEG as the source signal.
	
		\begin{table*}[width=\linewidth,cols=4,pos=h]
			\centering
			\caption{Summary of EEG-based BCI systems.}
			\label{tab:EEGBCI}
				\begin{tiny}
					\begin{tabular}{p{1in}p{1.3in}p{2.5in}p{1.4in}}
						\toprule
						Publication & Event under investigation & Implementation details&Dataset\\
						\midrule
						\citet{obermaier_information_2001}, \citeyear{obermaier_information_2001} & 5 tasks BCI system (imagining left-hand, right-hand, foot, tongue movements, or simple calculation). & A 5 state HMM with 8 (max) Gaussian mixtures per state, was used to model the spatiotemporal patterns in each signal segment. Features were extracted from all electrodes and fused into a combined feature vector and it had its dimensionality reduced before use in building the model. The expectation-maximization algorithm was used for the estimation of the transition matrix and the mixtures. & Data from 3 male subjects were collected for motor imagery tasks with the participants free of any medical or central nervous system conditions.\\
						\hline
						\citet{obermaier_hidden_2001}, \citeyear{obermaier_hidden_2001} & Two class motor imagery (left and right hands) BCI & Two 5 state HMMs (one for each class) with 8 (max) Gaussian mixtures per state, was used to model the spatiotemporal patterns in each signal segment. The Hjorth parameters of two channels (C3 and C4) were fused and fed into the HMM models to calculate the single best path probabilities for both models. The expectation-maximization algorithm was used for the estimation of the transition matrix and the mixtures. & Data from 4 male subjects were collected for motor imagery tasks with the participants free of any medical or central nervous system conditions.\\
						\hline
						\citet{pfurtscheller_graz-bci_2003}, \citeyear{pfurtscheller_graz-bci_2003} & Two class motor imagery BCI for virtual keyboard control & Two HMMs, one for each class, were trained and the maximal probability achieved by the respective HMM-model represents the chosen class. & Signals from two bipolar channels were acquired from three able-bodied subjects.\\
						\hline
						\citet{solhjoo_classification_2005}, \citeyear{solhjoo_classification_2005} & EEG-based mental task classification (left or right hand movement) & Discrete HMM and multi-Gaussian HMM -based classifiers have been used for raw EEG signals. & Dataset III of BCI Competition II (2003) provided by the BCI research group at Graz University \cite{pfurtscheller_dataset_2003}.\\
						\hline
						\citet{suk_two-layer_2010}, \citeyear{suk_two-layer_2010} & Multi-class motor imagery classification & In this study, dynamic patterns in EEG signals were modeled using two layers HMM. First time-domain patterns were extracted from the signals and have dimension reduced using PCA. Second, the likelihood for each channel is computed in the first layer of HMM and assembled in vector whose dimension is reduced with PCA as well. finally, the class label is calculated through the largest likelihood in the upper layer of HMM. Baum-Welch algorithm was used to estimate the parameters of the initial state distribution, the state transition probability distribution, and the observation probability distribution and Viterbi algorithm was used for decoding the state sequence. & Dataset IIa of BCI Competition IV (2008) provided by the BCI research group at Graz University \cite{brunner_dataset_2008}.\\
						\hline
						\citet{speier_integrating_2014}, \citeyear{speier_integrating_2014} & P300 speller & An HMM was used to model typing as a sequential process where each character selection is influenced by previous selections. The Viterbi algorithm was used to decode the optimal sequence of target characters. & Data were collected from 15 healthy graduate students and faculty with normal or corrected to normal vision between the ages of 20 and 35.\\
						\hline
						\citet{erfanian_real-time_2005}, \citeyear{erfanian_real-time_2005} & Real-time adaptive noise canceler for ocular artifact suppression in EEG & A recurrent multi-layer perceptron with a single hidden layer was trained for the noise canceling with the inputs as the contaminated EEG signal and the reference EOG. & A simulated EEG dataset was used for this study, generated through Gaussian white noise-based autoregressive process.\\
						\hline
						\citet{forney_classification_2011}, \citeyear{forney_classification_2011} & EEG signal forecasting and mental tasks classification & An Elman RNN was trained for forecasting EEG a single time step ahead then an Elman RNN-based classifier was trained to classify the mental task associated with the EEG signals. & 4 class dataset was collected from 3 subjects including combinations of the following mental tasks: clenching of right hand, shaking of left leg, visualization of a tumbling cube, counting backward from 100 by 3's, and singing a favorite song.\\
						\hline
						\citet{balderas_alternative_2015}, \citeyear{balderas_alternative_2015} & EEG classification for 2 class motor imagery (left hand and right hand) & An LSTM based classifier was trained and evaluated for EEG oscillatory components classification and compared with the regular neural network implementations. & BCI competition IV (2007) dataset 2b \cite{leeb_brain-computer_2007}\\
						\hline
						\citet{maddula_deep_2017}, \citeyear{maddula_deep_2017} & P300 BCI classification & A 3D CNN in conjunction with a 2D CNN were combined with an LSTM-based RNN to capture spatio-temporal patterns in EEG. & Data from P300 segment speller were collected, where the subjects mentally noted whenever the flashed letter is part of their target \cite{stivers_spelling_2017}. \\
						\hline
						\citet{thomas_deep_2017}, \citeyear{thomas_deep_2017} & Steady-state visual evoked potential (SSVEP)-based BCI classification & A single layer BRNN was used to perform classification and compared to different architecture and traditional classifying techniques. & 5-class SSVEP dataset \cite{oikonomou_comparative_2016}.\\
						\hline
						\citet{spampinato_deep_2017}, \citeyear{spampinato_deep_2017} & Visual object classifier using EEG signals evoked by visual stimuli & An LSTM based encoder to learn high order and temporal feature representations from EEG signals and then a classifier is used for identifying the visual object tat generated the stimuli. The authors here tested different architectures for the encoder including a common LSTM for all channels, channel LSTMs + common LSTM, and Common LSTM + fully connected layer. The authors also trained a CNN-based regressor for generating the EEG features to replace the whole EEG module and work only using source images of visual stimuli. & A subset of ImageNet dataset (40 classes) \cite{russakovsky_imagenet_2015} was used to generate visual stimuli for six subjects while EEG data is recorded.\\
						\hline
						\citet{hosman_bci_2019}, \citeyear{hosman_bci_2019} & Intercortical BCI for cursor control & An single layer LSTM-based decoder was built with three outputs to generate the cursor speed in x and y directions in addition to the distanc to target. & Intercortical neural signals recorded from three participants, each with 2 96-channel micro-electrode arrays \cite{hochberg_neuronal_2006}.\\
						\hline
						\citet{zhang_making_2020}, \citeyear{zhang_making_2020} & EEG-Based Human Intention Recognition & In this study, multi-channel raw EEG sequences into mesh-like representations that can capture spatiotemporal characteristics of EEG and its acquisition. These meshes are then fed into deep neural networks that perform the recognition process. Multiple network architectures were investigated including a CRNN that starts with a 2D CNN that processes the meshes followed by a two-layer LSTM-based RNN to extract the temporal features, then a fully connected layer and an output layer. The second network investigated was composed of two parallel branches the first was a two layer LSTM-based RNN to extract the temporal features and the second was a multi-layer 2D/3D CNN to extract the spatial features and the output from the two branches is fused and used for recognition. This study used fusion on both data-level and feature-level. & EEG Motor Movement/Imagery Dataset \cite{schalk_bci2000_2004,goldberger_physiobank_2000}. \\
						\hline
						\citet{tortora_deep_2020}, \citeyear{tortora_deep_2020} & BCI for gait decoding from EEG & EEG data were preprocessed to remove motion artifacts through high pass filtration and independent component analysis. Different frequency bands were then extracted and a separate classifier is trained based on each frequency band. The classifiers were based on a two-layer LSTM-based RNN followed by a fully connected layer, a softmax layer, and an output layer that manifests the prediction output.  & EEG data were recorded from 11 subjects walking on a treadmill using a 64-channel amplifier and 10/20 montage.\\
						\bottomrule
					\end{tabular}
				\end{tiny}
		\end{table*}

\section{Event detection in EMG}
	Electromyography (EMG) is the method of sensing the electric potential evoked by the activity of muscle fibers as driven by the spikes from spinal motor neurons. EMGs are recorded either using surface electrodes or via needle electrodes; however, surface EMG (sEMG) is rarely used clinically in the evaluation of neuromuscular function and its use is limited to the measurement of voluntary muscle activity \cite{zwarts_multichannel_2003}. Routine evaluation of the neuromuscular function is typically performed using needle (invasive) EMG that, despite of its effectiveness and the availability of several electrode types that suite many clinical questions, is often painful and traumatic and may lead to the destruction of several muscle fibers \cite{zwarts_multichannel_2003,hogrel_clinical_2005}. sEMG has been widely used as control signals for multiple applications especially in rehabilitation including but not limited to body-powered prostheses, grasping control, and gesture based interfaces \cite{oskoei_myoelectric_2007}. A myoelectric signal usually has its manifested events as two states, the first is the transient state which emanates as the muscle goes from the resting state to voluntary contraction. The second is the steady state which represents maintaining the contraction level in the muscle \cite{oskoei_myoelectric_2007}. It has been shown that the steady state segments are more robust as control signals compared to the transient state due to longer duration and better classification rates \cite{englehart_wavelet-based_2001}. As follows in Table \ref{tab:EMGSummary}, we give a review about the recent advances in the detection of myoelectric events in EMG signals.

	\begin{table*}[width=\linewidth,cols=4,pos=h]
		\centering
		\caption{Summary of event detection in EMG signals.}
		\label{tab:EMGSummary}
			\begin{tiny}
				\begin{tabular}{p{1in}p{1.3in}p{2.5in}p{1.4in}}
					\toprule
					Publication & Event under investigation & Implementation details&Dataset\\
					\midrule
					\citet{chan_continuous_2005}, \citeyear{chan_continuous_2005} & Continuous identification of six classes hand movement in sEMG & An HMM with uniformly distributed initial states and Gaussian observation probability density function whose parameters can be completely estimated from the training data, was constructed for the detection process. The expectation-maximization algorithm wasn't used here due to the assumption of uniform initial state probabilities and directly estimating the Gaussian parameters from the training data. Overlapping 256 ms observation windows were used and in each observation window the root mean square value and the first 6 autoregressive coefficients were computed as features. & 4-channel sEMG collected from the forearm of 11 subjects for six distinct motions (wrist flexion, wrist extension, supination, pronation, hand open, and hand close) \cite{englehart_continuous_2003}. \\
					\hline
					\citet{zhang_framework_2011}, \citeyear{zhang_framework_2011} & Hand gesture recognition in acceleration and sEMG & In this work, the authors actually identified the active segments via processing and thresholding of the average signal of the multichannel sEMG. The onset is when the energy is higher than a certain threshold and the offset when the energy is lower than another threshold. Features from time, frequency, and time-frequency domains were extracted from both acceleration signals and sEMG, and fed to five-state HMMs for classification. Baum-Welch algorithm was used for training with Gaussian multivariate distribution for observations. Decision making here is done in a tree-structure (decision-level fusion) through four layers of classifiers with the last layer as the HMM. & sEMG and 3d acceleration were collected from two right-handed subjects who performed 72 Chinese sign language words in a sequence with 12 repetitions per motion, and a predefined 40 sentences with 2 repetitions per sentence. \\
					\hline
					\citet{wheeler_gesture-based_2006}, \citeyear{wheeler_gesture-based_2006} & Hand gesture recognition in sEMG & Moving average was used on the sEMG signals to provide the input for continuous left-to-right HMMs with tied Gaussian mixtures. The training was performed using the Baum-Welch algorithm and the real-time recall was performed with The Viterbi algorithm. The models were also initialized using K-means clustering so that the states were partitioned to equalize the amount of variance within each state. This study employed feature-level fusion to combine multi-channel data. & Data from one participant repeating 4 gestures on a joystick (left, right, up, and down) for 50 times per gesture, were collected using four pairs of dry electrodes. Another portion of data was collected using 8 pairs of wet electrodes on gestures of typing on a number pad keyboard (0-9) for 40 strokes on each key.\\
					\hline
					\citet{monsifrot_sequential_2014}, \citeyear{monsifrot_sequential_2014} & Extraction of the activity of individual motor neurons in single channel intramuscular EMG (iEMG) & The iEMG signal was modeled as a sum of independent filtered spike trains embedded in noise. A Markov model of sparse signals was introduced where the sparsity of the trains was exploited through modeling the time between spikes as discrete weibull distribution. An online estimation method for the weibull distribution parameters was introduced as well as an implementation of the impulse responses of the model. & The method introduced was tested over both simulated and experimental iEMG signals. the simulated signals were generated via Markov model under 10 kHz sampling frequency and with filter shapes obtained from experimental iEMG for more realistic simulation. The experimental iEMG signals were acquired from the extensor digitorum of a healthy subject with teflon coated stainless steel wire electrodes.\\
					\hline
					\citet{lee_emg-based_2008}, \citeyear{lee_emg-based_2008} & sEMG-based speech recognition & A continuous HMM was constructed with Gaussian mixtures model adopted for sEMG-based word recognition based on log mel-filter bank spectrogram of the windowed EMG signals. The segmental K-means algorithm was used for optimal HMM parameters estimation where HMM parameters for the i\textsuperscript{th} state and k\textsuperscript{th} word are estimated from the observations of the corresponding state of the same word. Viterbi algorithm was used for the decoding process. & EMG signals were collected from articulatory facial muscles from 8 Korean male subjects. The subjects were asked to pronounce each word from a 60-word vocabulary in a consistent manner in addition to generating a random set of words based on this vocabulary. \\
					\hline
					\citet{chan_hidden_2002}, \citeyear{chan_hidden_2002} & sEMG-based automatic speech recognition & A six state left-right HMM with single mixture observation densities, was constructed for identifying the words based on three features extracted from sEMG that included the first two autoregressive coefficients and the integrated absolute value. HMM was trained in this work using the expectation-maximization algorithm.  & sEMG from five articulatory facial muscles were collected. The dataset used here was a subset of the dataset described in \cite{chan_myo-electric_2001} with ten-English word vocabulary.\\
					\hline
					\citet{li_muscle_2014}, \citeyear{li_muscle_2014} & Identification/prediction of functional electrical stimulation (FES)-induced muscular dynamics with evoked EMG (eEMG) & A nonlinear ARX-type RNN was used to predict the stimulated muscular torque and track muscle fatigue. The model takes the eEMG as an input and produces the predicted torque. & The experiments were conducted on 5 subjects with spinal cord injuries.\\
					\hline
					\citet{xia_emg-based_2018}, \citeyear{xia_emg-based_2018} & Hand motion estimation from sEMG & A CRNN with 3 CNN layers and 2 LSTM layers was used for the prediction and the model used the power spectral density as input.  & sEMG signals were collected from 8 healthy subjects using 5 pairs of bipolar electrodes placed on shoulder to record EMG from biceps brachii, triceps brachii, anterior deltoid, posterior deltoid, and middle deltoid. The hand position in 3D space was tracked as the objective for this system. \\
					\hline
					\citet{quivira_translating_2018}, \citeyear{quivira_translating_2018} & Simple hand finger movement identification in sEMG & An LSTM-based RNN was used to implement a recurrent mixture density network (RMDN) \cite{graves_generating_2013} that probabilistically model the output of the Network in order to capture the complex features present the hand movement.& 8 channel EMG signals were collected from the proximal forearm region, targeting most muscles used in hand manipulation. The hand pose tracking was performed with a Leap Motion sensor and the subjects were asked to perform 7 hand gestures with repetitions per gesture. \\
					\hline
					\citet{hu_novel_2018}, \citeyear{hu_novel_2018} & Hand gesture recognition in sEMG & sEMG signals from all channels were segmented into windows of fixed size and transformed into an image representation that was then fed into a CNN with two convolutional layers, two locally connected layers, and three fully connected layers followed by an LSTM-based RNN and an attention layer to enhance the output of the network. & Experiments were performed over the first and second sub-databases of NinaPro (Non Invasive Adaptive Prosthetics) database \cite{atzori_electromyography_2014}.\\
					\hline
					\citet{samadani_gated_2018}, \citeyear{samadani_gated_2018} & EMG-Based Hand Gesture Classification & Different RNN architectures were tested in this study to chose the best performing architecture. The evaluated models included uni and bidirectional LSTM- and GRU-based RNNs with attention mechanisms. The models worked on the preprocessed (denoised) raw EMG signals.  & Publicly-available NinaPro hand gesture dataset (NinaPro2) was used \cite{atzori_building_2012}.\\
					\hline
					\citet{simao_emg-based_2019}, \citeyear{simao_emg-based_2019} & EMG-based online gestures classification & Features were extracted from multi-channel EMG (standard deviation along each time frame) and fed into a dynamic RNN model that is composed of a dense layer followed by an LSTM-based RNN layer and another dense layer followed by the output layer. This model was compared to a similar GRU-based model and another static feed forward neural network model. This study used combined feature vector as an input for the models. & the synthetic sequences of the UC2018 DualMyo dataset \cite{simao_uc2018_2018} and a similar subset of the NinaPro DB5 dataset \cite{pizzolato_comparison_2017} \\
					\bottomrule
				\end{tabular}
			\end{tiny}
	\end{table*}

\section{Event detection in other biomedical signals}
	Physiological monitoring is an essential part of all care units nowadays and it is not limited to the aforementioned biomedical signals only. Tens of variables are collected in the form of time series containing hundreds of events that are of importance to the diagnosis and treatment/rehabilitation. Event detection methods have had a strong presence in the analysis of such series. For instance, cardiovascular disorders are not only assessed through ECG but also phonocardiogram is used as an easier way for general practitioner to identify the changes in heart sounds. Extracting the cardiac cycle has been one of the major problems in phonocardiogram as well and was addresses using HMMs in multiple pieces of work \cite{schmidt_segmentation_2010,ricke_automatic_2005,sedighian_pediatric_2014,lima_automatic_2008}. On the other hand, most of RNN based methods in phonocardiogram, have been used for pure classification purposes and anomaly recognition \cite{thomas_deep_2017}.
	
	3D acceleration is an emerging technology as well, that has been extensively used in the assessment and detection of many medical conditions in swallowing \cite{sejdic_computational_2019} and human gait analysis \cite{shull_quantified_2014}. In swallowing, acceleration signals have been used for the detection of pharyngeal swallowing activity via maximum likelihood methods with minimum description length in \cite{sejdic_segmentation_2009} and using short time Fourier transform and neural networks in \cite{khalifa_non-invasive_2020}. RNNs were also employed for event detection in swallowing acceleration signals including the upper esophageal sphincter opening in \cite{khalifa_upper_2020,donohue_how_2020}, laryngeal vestibule closure \cite{mao_estimation_2020}, and hyoid bone motion during swallowing \cite{mao_neck_2019}. In gait analysis, HMMs were used for recognition and extraction in multiple occasions \cite{nickel_using_2011,mannini_hidden_2011,nickel_classifying_2013,panahandeh_continuous_2013} as well as RNNs \cite{inoue_deep_2018,lisowska_evaluation_2015,theodoridis_human_2017}. 

\section{Challenges and Future Directions}
	Event detection in biomedical signals is a critical step for diagnosis and intervention procedures that are extensively used on a daily basis in nearly every standard clinical setting. It also represents the core of various eHealth technologies that employ wearable devices and regular monitoring of physiological signs. Being such a fundamental operation that controls the clinical decision making process, it necessitates precise detection in a fairly complex environment that contains multiple events occurring concurrently. Particularly, false positive rate in clinical testing is an important indicator for how well the detection model generalizes and differentiates between the event of interest and the background noise. Building such highly accurate models depends on many factors that include the diversity in the used dataset and labels in addition to model capacity. 
		
	\subsection{Classical Models Scaling: Challenges}
		As mentioned before, biomedical signals are the manifestation of well-coordinated, yet complex physiological processes which involve various anatomical structures that are close in position and share several functions. Hence, the collected signals pick not only the target physiological process but also other unavoidable neighbor processes. An example of that is the detection of the combined activation for multiple muscles in sEMG, eye blinking along with neural activity in EEG, and head movement along with swallowing vibrations in swallowing accelerometry. Extraction of the event of interest in this case requires the exhausting labeling of the underlying set of processes in order to be able to build the predefined state space for classical stochastic methods such as HMM, from which the state sequence is drawn. Manual labeling or interpretation of the biomedical signals is not only an exhausting task, but also requires extensive domain knowledge and expertise to perform.\par
		
		One way that can be used to enhance the expressive power of stochastic models such as HMM, is the inclusion of non-Gaussian mixtures which can boost the performance in many cases because Gaussianity is not always a reasonable assumption in many applications. One of the mixtures that was proposed as an extension for non-Gaussian mixtures, is independent component analyzers mixture model (ICAMM) and it has been applied in multiple biomedical signal applications such as sleep disorders detection and classification of neuropsychological tasks in EEG \cite{safont_multichannel_2019, salazar_including_2010}.
		
		An additional way to increase the model capacity and its ability to model the underlying sequence of events, is through using strongly representing domain features. One of the most popular domains representations, is wavelet decomposition which has proven its superiority to provide high level representation of events in a wide variety of biomedical signals such as phonocardiograms \cite{lima_automatic_2008,huiying_heart_1997}, EEG \cite{abdullah_seizure_2012,petrosian_recurrent_2000,kumar_automated_2008,minasyan_patient-specific_2010}, and EMG \cite{englehart_wavelet-based_2001}. Handcrafting features, however, is not an easy task and requires an extensive domain knowledge and significant efforts to come up with cues that trigger the identification of specific signal components. Furthermore, mapping the feature space into a more comprehensive space of less dimensionality is often a paramount operation prior to building the model. Given the previous factors, models that are able to learn high level representations simultaneously from raw signals and have the massive expressive power to model tasks involving long time lags, can be of a great benefit \cite{lecun_gradient-based_1998}. 
	
	\subsection{High Capacity Models Embedding Feature Extraction}
		The evolution of deep learning has revolutionized the way in which problems are addressed and instead of classification and detection systems that solely relied on handcrafted features, end-to-end systems are being trained to take care of all steps from the raw input till the final output. End-to-end systems are complex, although rich, processing pipelines that make the most of the available information through using a unified scheme that trains the system as a whole from the input till the output is produced \cite{glasmachers_limits_2017}. It has been shown that deep architectures can replace handcrafted feature extraction stages and work directly on raw data to produce high levels of abstraction. RNNs have been introduced in \citeyear{cheron_dynamic_1996} for the identification of arm kinematics during hand drawing from raw EMG signals \cite{cheron_dynamic_1996} and then the same architecture was adopted for lower limb kinematics in \cite{cheron_dynamic_2003}. In both studies, the authors verified that an RNN was able to map the relationship between raw EMG signals and limbs' kinematics during drawing for the arm and human locomotion for the lower limb. \citet{chauhan_anomaly_2015} and \citet{sujadevi_real-time_2017} have also used more sophisticated multi-layer LSTM-based RNN architectures on raw ECG signals for arrhythmia detection. \citet{spampinato_deep_2017} have employed RNNs as well to extract discriminative brain manifold for visual categories from EEG signals. Further, \citet{vidyaratne_deep_2016} used RNNs for seizure detection in EEG; however, they used a denoised and segmented version of the signals. As mentioned earlier, although RNNs are efficient in modeling long contexts, they tend to have the error signals propagate through a tremendous number of steps when being fed highly sampled inputs such as raw signals which affects the network optimizability and training speed \cite{schwab_beat_2017,stollenga_parallel_2015}.
		
		In this regard, convolutional neural networks (CNNs) have been utilized to perceive small local contexts which then are propagated to an RNN for the perception of temporal contexts or a feed-forward network for a classification or prediction target. CNNs were introduced as a solution to enable recognition systems to learn hierarchical internal representations that form the scenes in vision applications (pixels form edglets, edglets form motifs, motifs form parts, parts form objects and objects form scenes) \cite{lecun_convolutional_2010,lecun_handwritten_1990}. Thus, CNNs are basically multi-stage trainable architectures that are stacked on top of each other to learn each level of the feature hierarchy \cite{lecun_gradient-based_1998,lecun_convolutional_2010}. Each stage is usually composed of three layers, a filter bank layer, a non-linear activation layer, and a pooling layer. A filter bank layer extracts particular features at all locations on the input. The non-linear activation works as a regulator that determines whether a neuron should fire or not through checking the its value and deciding if the following connections should consider this neron activated \cite{lecun_gradient-based_1998}. A pooling layer represents a dimensionality reduction procedure that processes the feature maps in order to produce lower resolution maps that are robust to the small variations in the location of features \cite{lecun_convolutional_2010}. The coefficients of the filters are the trainable parameters in the CNNs and they are updated simultaneously by the training algorithm to minimize the discrepancy between the actual output and the desired output \cite{lecun_convolutional_2010}.
		
		The design concept of CNNs first evolved for vision applications; but since then, the same concept is being adopted for pattern analysis and recognition in biomedical signals \cite{cecotti_convolutional_2011,kiranyaz_real-time_2016,shashikumar_detection_2018,tan_application_2018,xiong_ecg_2018,xia_emg-based_2018}. For instance, \citet{shashikumar_detection_2018} used a 5-layer 2D CNN followed by a BRNN in association with soft attention mechanism to process the wavelet transform of ECG signals for the detection of atrial fibrillation. \citet{tan_application_2018} also used a 1D 2-layer CNN with a 3-layer LSTM-based RNN for the detection of coronary artery disease in ECG. Further, \citet{xiong_ecg_2018} used a residual convolutional recurrent neural network for the detection of cardiac arrhythmia in ECG. All these experiments using RNNs on top of CNNs for biomedical signal analysis were successful to produce extremely high levels of abstraction and rich temporal representation that can perceive long range contexts without human intervention in addition to being easier to optimize computationally. CNNs have been also utilized in association with fully connected networks to increase the capacity of HMMs in connectionist hybrid DNN-HMM models due to the ability of CNNs to process high-dimensional multi-step inputs \cite{saidutta_increasing_2018}. Such hybrid systems provided state of the art performance especially in the field of handwriting recognition \cite{wang_comprehensive_2018,wang_writer-aware_2020}.
	
	\subsection{Transfer Learning}
		Despite the fact that most of the previously mentioned methods are achieving great results on certain datasets, it is popular that they can easily overfit the data, resulting in poor generalization. Thus, it requires not only very large but also diverse datasets to train and validate models that well generalize. In biomedical signal processing field, the collection of such datasets may pose a challenge towards developing reliable models. Strictly speaking, it may not be feasible to find a large population of subjects when studying a rare disease and yet if it is feasible, it is extremely difficult to acquire the expert reference annotations for the underlying dataset \cite{dvornek_learning_2018}. Many factors contribute to this, as mentioned before, the noisy nature of biomedical signals increases the difficulty of manual interpretation and necessitates the presence of reference modalities to acquire accurate information about the processes such as collection of x-ray videofluoroscopy simultaneously with swallowing accelerometry \cite{yu_silent_2019}. Another factor is that the experts annotating the data need to maintain high record of reliability across time and to be compared to peer experts which might be difficult to achieve or require continuous training and checking of the experts' reliability.
		
		One way to overcome limited- size and/or diversity datasets, is to utilize the the pretrained models from relatively different domains and apply them to solve the particular targeted problem or so-called transfer learning \cite{pan_survey_2010}. In transfer learning, the pretrained model's weights are used as initialization and then fine-tuned accordingly to fit the new dataset. In most cases, retraining happens in a much lower (~10 times smaller) learning rate than the original. Transfer learning has been used for event detection and classification tasks in multiple biomedical signals including ECG for cardiac arrhythmia detection \cite{isin_cardiac_2017}, EEG for drowsiness detection \cite{wei_selective_2015} and driving fatigue detection \cite{zhang_transfer_2015}, and EMG for hand gesture classification \cite{cote-allard_deep_2019}. However, one thing worth mentioning is that transfer learning sometimes may not help perform better than the originally trained model if there exist huge differences between the datasets or deterioration in inter-subject variability \cite{pan_survey_2010}.

\section{Conclusion}
	In this paper, we provided a comprehensive review of event extraction methods in biomedical signals, in particular hidden Markov models and recurrent neural networks. HMM is a probabilistic model that represents a sequence of observations in terms of a hidden sequence of states and sets the concepts and methods on how to find the optimal state sequence that best describes the observations. RNN is a type of neural networks that was introduced to model the time dependency and perform contextual mapping in sequences. This review showed that the presence of dynamic programming algorithms like the EM and Viterbi, led to the wide spread of HMMs which were used to dynamically transcribe the context of many biomedical signals. It wasn't too long until HMMs became insufficient for time series modeling needs, specifically modeling long range dependencies and larger state spaces, and RNNs started to gradually replace HMMs in time-dependent contextual mappings. So far, RNNs have proven superiority in time series modeling especially in biomedical signals and continue to expand their domination in building automatic detection and diagnosis systems through the emerging designs and practices experimented in nearly every field.

\section*{Acknowledgments}
The work reported in this manuscript was supported by the National Science Foundation under the CAREER Award Number 1652203. The content is solely the responsibility of the authors and does not necessarily represent the official views of the National Science Foundation.

\bibliographystyle{model1-num-names}

\end{document}